\documentclass[lettersize,journal]{IEEEtran}
\usepackage{amsmath,amsfonts}
\usepackage{amssymb}
\usepackage{array}
\usepackage[caption=false,font=normalsize,labelfont=sf,textfont=sf]{subfig}
\usepackage{textcomp}
\usepackage{stfloats}
\usepackage{url}
\usepackage{verbatim}
\usepackage{graphicx}
\usepackage{cite}
\usepackage{graphicx}
\usepackage{amsmath,amsfonts}
\usepackage{algpseudocode}
\usepackage{enumitem}
\usepackage{wrapfig}
\usepackage{subcaption}
\usepackage{multirow}
\usepackage{makecell} 
\usepackage{tabularx}
\usepackage{colortbl}
\usepackage[table]{xcolor}
\usepackage{adjustbox}
\usepackage{pifont}
\usepackage{array}
\usepackage{float}
\usepackage{booktabs}
\usepackage{hyperref}
\usepackage[ruled,vlined]{algorithm2e}
\begin{document}

\title{Aligning MLLM Benchmark With Human Preferences via Structural Equation Modeling}

\author{Shengwu Xiong, Tianyu Zou, Cong Wang,~\IEEEmembership{Member,~IEEE}, Xuelong Li,~\IEEEmembership{Fellow,~IEEE}
\thanks{This work was supported in part by the National Key Research and Development Program of China under Grant No. 2022ZD0160604, in part of the National Natural Science Foundation of China under Grant 62476219, in part by the National Key R\&D Program of Shanxi under Grant 2024CY2-GJHX-54, in part by the Young Talent Fund of Association for Science and Technology in Shaanxi, China under Grant 20230140, and in part by the Fundamental Funds for the Central Universities. \textit{(Corresponding authors: Cong Wang.)}}
\thanks{
Shengwu Xiong is with the Interdisciplinary Artificial Intelligence Research Institute, Wuhan College, Wuhan 430212, China, also with School of Computer and Artificial Intelligence, Wuhan University of Technology, Wuhan 430070, China, and also with Shanghai Artificial Intelligence Laboratory, Shanghai 200232, China.
Tianyu Zou is with the School of Computer and Artificial Intelligence, Wuhan University of Technology, Wuhan 430070, China, also with Sanya Science and Education Innovation Park, Wuhan University of Technology, Sanya 572000, China, and also with the Institute of Automation, Chinese Academy of Sciences, Beijing 100190, China.
Cong Wang is with the School of Mathematics and Statistics, Northwestern Polytechnical University, Xi'an 710129, China, and also with Shanghai Artificial Intelligence Laboratory, Shanghai 200232, China.
Xuelong Li is with the Institute of Artificial Intelligence (TeleAI) of China Telecom and also with Shanghai Artificial Intelligence Laboratory, Shanghai 200232, China. Email: xiongsw@whut.edu.cn; zoutianyu@whut.edu.cn; congwang0705@nwpu.edu.cn; xuelong li@ieee.org.}}


\markboth{Journal of \LaTeX\ Class Files,~Vol.~14, No.~8, August~2021}%
{Shell \MakeLowercase{\textit{et al.}}: A Sample Article Using IEEEtran.cls for IEEE Journals}


\maketitle

\begin{abstract}
Evaluating multimodal large language models (MLLMs) remains a fundamental challenge due to a lack of structured, interpretable, and theoretically grounded benchmark designs. Existing benchmarks often adopt heuristic-based task groupings with unclear cognitive targets, thus resulting in overlapping abilities, redundant indicators, and limited diagnostic power. To do as, we propose a novel framework for aligning MLLM benchmark based on structural equation modeling to analyze and quantify internal validity, dimensional separability, and contribution of benchmark components. Motivated by the observed limitations of current designs, we further introduce a novel capability hierarchy grounded in Piaget’s theory of cognitive development, dividing MLLM abilities into three hierarchical layers, i.e., Perception, Memory, and Reasoning. We reorganize existing MLLM benchmarks under the proposed framework and construct a new benchmark named \textsc{Gold}. Experimental results demonstrate that the proposed benchmark exhibits stronger interpretability, reduced indicator redundancy, and clearer cognitive consistency compared to existing approaches. Our benchmark is available at \url{https://github.com/tianyu-zou/GoldBench}.
\end{abstract}


\section{Introduction}
\IEEEPARstart{T}{he} rapid advancements in the field of multimodal learning have been driven by the emergence of increasingly powerful and versatile Multimodal Large Language Models (MLLMs)~\cite{kuang2025natural, qin2024multilingual, zhang2024mm}. From OpenAI’s GPT-4o~\cite{achiam2023gpt} and Google’s Gemini~\cite{team2023gemini} to a variety of open-source models such as LLaVA~\cite{liu2023visual, touvron2023llama}, mPLUG-Owl~\cite{ye2023mplug}, and BLIP-2~\cite{li2023blip}, these models exhibit remarkable capabilities in integrating and processing multiple modalities, including text, images, and videos. As MLLMs continue to evolve~\cite{zhang2025zooprobe, niu2024text}, the need for comprehensive evaluation frameworks becomes increasingly critical to assess their reasoning abilities, multimodal understanding, and generalization performance~\cite{fu2024mme, li2024embodied}. A rigorous evaluation is essential not only for benchmarking advancements in model development but also for identifying potential biases, addressing ethical concerns, and ensuring the responsible and reliable deployment of these models across various real-world applications~\cite{feng2025benchmarking, qin2025survey, zhang2024q}.

\begin{figure}
  \centering
  \includegraphics[width=1\linewidth]{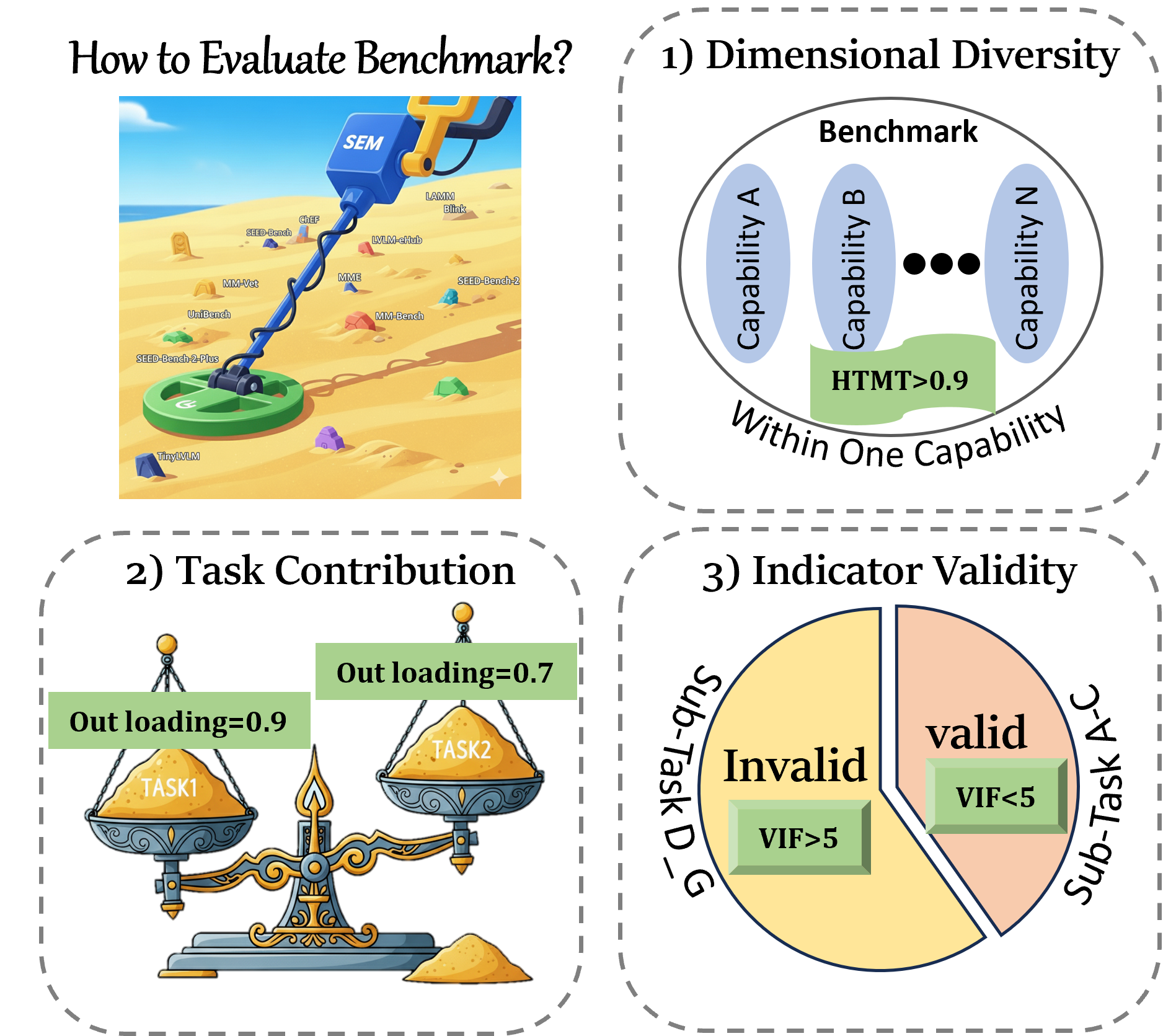}
  \caption{Brief illustrations of How to Evaluate Benchmark?}
  \label{fig:motivation}
\end{figure}

Evaluations not only reveal the advantages and shortcomings of these models across various tasks and dimensions, but they also provide clear directions for future enhancements~\cite{zhu2023large, li2024survey, huang2024survey}. Currently, the evaluation of MLLMs primarily relies on benchmark datasets composed of large-scale question-answer pairs~\cite{guo2025r, li2025benchmark}. According to statistics from Opencompass~\cite{2023opencompass}, there are now over 300 evaluation benchmarks for MLLMs, with more than 100 of them integrated into various multimodal evaluation suites. However, the rapid proliferation of evaluation benchmarks has also introduced new challenges: the lack of unified standards makes it difficult to assess the quality, coverage, and reliability of these benchmarks~\cite{li2025information, zhang2025redundancy}. As a result, it becomes increasingly hard to determine which benchmarks offer meaningful insights into model capabilities and which may lead to biased or misleading conclusions. As shown in Fig.~\ref{fig:motivation}, we develop a structural equation modeling~\cite{ullman2012structural} based framework to evaluate the quality and contribution of MLLM benchmarks, where SEM serves as a statistical approach to link observed task performances with latent capability dimensions, enabling us to evaluate the quality and contribution of MLLM benchmarks.

\begin{table}[h]
    \centering
    \caption{Comprehensive Benchmark Overview}
    \resizebox{\columnwidth}{!}{
    \begin{tabular}{lccc}
    \toprule
        Benchmark & \makecell{Capability \\ Dimension} & \makecell{Task \\ Number} & \makecell{Publication \\ Date} \\ \midrule
        MME~\cite{yin2024survey} & 4 & 14 & Jun-23 \\
        LVLM-eHub~\cite{xu2024lvlm} & 6 & 16 & Jun-23 \\
        SEED-Bench~\cite{li2023seed} & 2 & 12 & Aug-23 \\
        TouchStone~\cite{bai2023touchstone} & 5 & 27 & Sep-23 \\
        LAMM~\cite{yin2023lamm} & 4 & 9 & Nov-23 \\
        ChEF~\cite{shi2023chef} & 6 & 9 & Nov-23 \\
        SEED-Bench-2~\cite{li2024seed} & 6 & 27 & Nov-23 \\
        SEED-Bench-2-Plus~\cite{li2024seed} & 3 & 63 & Apr-24 \\
        MMT-Bench~\cite{ying2024mmt} & 32 & 162 & Apr-24 \\
        BLINK~\cite{fu2024blink} & 3 & 14 & Jul-24 \\
        TinyLVLM~\cite{shao2025tinylvlm} & 6 & 42 & Aug-24 \\ 
        UniBench~\cite{al2024unibench} & 17 & 53 & Aug-24 \\ 
        MMStar~\cite{chen2024we} & 6 & 18 & Sep-24 \\
        MMBench~\cite{liu2024mmbench} & 6 & 20 & Sep-24 \\
        MM-Vet~\cite{yu2023mm} & 6 & 16 & Dec-24 \\       
        \bottomrule
    \end{tabular}}
    \label{tab: Comprehensive_benchmark}
\end{table}

Recent representative multi-modal benchmarks have shown a trend toward the widespread adoption of tiered capability frameworks, which aim to deconstruct model performance into structured dimensions. As illustrated in Table~\ref{tab: Comprehensive_benchmark}, there is a clear pattern of increasing the number of tasks and capability dimensions across recent benchmarks. This trend is particularly evident within the same benchmark series. For instance, SEED-Bench~\cite{li2024seed} initially proposes a two-dimensional framework with 12 tasks, but its updated version, SEED-Bench-2~\cite{li2024seed}, expands to six dimensions and 27 tasks. Similarly, other benchmarks also exhibit this ``dimension expansion'', such as MVBench~\cite{li2024mvbench}, which divides 47 tasks based on six core capabilities, and MMTBench~\cite{ying2024mmt}, which introduces 32 dimensions and 162 fine-grained tasks. While this ongoing expansion reflects a deeper exploration of model understanding, it also raises an important methodological issue: How can we construct a theoretically grounded and empirically verifiable capability structure that avoids arbitrary task proliferation and ensures interpretability?

To address this issue, we develop a novel SEM-based pruning framework to define and validate the latent structure of model abilities based on observed task performance. While human evaluation is widely regarded as the gold standard for assessing reliability, its high labor cost and limited scalability make it impractical for large-scale benchmark analysis~\cite{liu2022revisiting}. This creates an urgent need for an alternative that can approximate human judgment while substantially reducing manual effort. Guided by human evaluation results~\cite{chiang2024chatbot, li2025humanpcr, awasthi2023humanely} as the reliability reference, our framework constructs SEM models to capture the latent structure of model abilities from observed task performance. In this process, SEM not only infers unobservable constructs such as perception and cognition, but also quantifies the causal contributions of individual tasks to the overall evaluation~\cite{lowry2014partial}. This enables the pruning of redundant tasks while retaining those most aligned with human judgment.

Building on this SEM-based pruning framework, we further ground the construction of the benchmark in Piaget's theory of cognitive development~\cite{piaget2000piaget}, using its account of how humans acquire perceptual, mnemonic, and reasoning abilities to derive the human-centered \textsc{Gold} benchmark. Traditional benchmarks, driven largely by heuristic task groupings, often yield inconsistent or overlapping ability hierarchies that neither reflect human cognition nor guide task design in a principled way~\cite{dimara2018task}. A Piaget‑inspired bottom‑up stratification offers a clear remedy. By aligning MLLM capabilities with the Sensorimotor, Preoperational, and (Concrete \& Formal) Operational stages of human development, we define three continuous ability layers in \textsc{Gold}: Perception, Memory, and Reasoning. This human‑cognitive foundation enhances interpretability and consistency in benchmarks.

Our main contributions are as follows:
\begin{itemize}
\setlength{\itemsep}{0.03em}
    \item \textbf{SEM-based Evaluation Framework:} To the best of our knowledge, we are the first to apply SEM in MLLM assessment, modeling the causal relationships between observable task metrics and latent capabilities. This approach not only introduces a novel evaluation pipeline but also ensures a transparent, verifiable, and reproducible framework for MLLM benchmark evaluation.
    \item \textbf{Piaget-Inspired Hierarchical Capability Structure:} By leveraging Piaget’s theory of cognitive development, we define a three-level ability hierarchy, Perception, Memory, and Reasoning, which systematically organizes tasks, mitigates dimension inflation, and enhances interpretability.
    \item \textbf{\textsc{Gold} Benchmark:} We propose the human-centered \textsc{Gold} benchmark, derived from our Piaget-inspired hierarchical framework, to evaluate MLLMs across perceptual, mnemonic, and reasoning tasks in a structured and cognitively grounded manner.
  \end{itemize}

\begin{table}[t]
\centering
\caption{Multimodal Benchmarks by Capability}
\resizebox{0.5\textwidth}{!}{
\begin{tabular}{lp{6.5cm}}
\toprule
\textbf{Capability} & \textbf{Benchmarks} \\ 
\midrule
Perception & Q-Bench~\cite{zhang2024q}, PEC~\cite{peng2024synthesize}, GVT-Bench~\cite{wang2023makes}, VBench~\cite{wu2024v}, OCRBench~\cite{liu2024ocrbench}, CODE~\cite{zang2025contextual}, MMUBench~\cite{wang2024mm}, MMVP~\cite{tong2024eyes}, CVBench~\cite{tong2024cambrian}, EQBEN~\cite{wang2023equivariant}, P$^2$GB~\cite{chen2024plug}, MDVP-Bench~\cite{lin2024draw}, MM-SAP~\cite{li2024single}, MagnifierBench~\cite{li2023otterhd}, UNIAA~\cite{zhou2024uniaa}, AesBench~\cite{huang2024aesbench}, II-Bench~\cite{liu2024ii}, ImplicitAVE~\cite{zou2024implicitave}, EmoBench~\cite{yang2024emollm}, FABA-Bench~\cite{li2024facial}, LaVy-Bench~\cite{tran2024lavy}, MMMB~\cite{sun2024parrot}, M3GIA~\cite{song2024m3gia}, SeaEval~\cite{wang2023seaeval}, CVQA~\cite{romero2024cvqa}, Henna~\cite{alwajih2024peacock}, MTVQA~\cite{tang2024mtvqa} \\[2pt]
Memory & CODIS~\cite{luo2024codis}, MMNeedle~\cite{wang2024multimodal}, MileBench~\cite{song2024milebench}, MM-NIAH~\cite{wang2024needle}, MuirBench~\cite{wang2024muirbench}, Mementos~\cite{wang2024mementos}, MMIU~\cite{meng2024mmiu}, Mantis-Eval~\cite{jiang2024mantis}, IIT~\cite{zhang2024wings}, VEGA~\cite{zhou2024vega}, MMC4~\cite{zhu2023multimodal}, Obelics~\cite{laurenccon2023obelics}, CoMM~\cite{chen2025comm}, VL-ICLBench~\cite{zong2024vl}, MMMU~\cite{yue2024mmmu}, VideoNIAH~\cite{zhao2024needle}, OSCaR~\cite{nguyen2024oscar}, TempCompass~\cite{liu2024tempcompass}, VITATECS~\cite{li2024vitatecs}, MovieChat-1K~\cite{song2024moviechat}, EgoSchema~\cite{mangalam2023egoschema}, TimeIT~\cite{ren2024timechat}, ADLMCQ~\cite{chakraborty2024llavidal}, MLVU~\cite{zhou2024mlvu}, Event-Bench~\cite{du2024towards}, WorldNet~\cite{ge2024worldgpt} \\[2pt]
Reasoning & SPIQA~\cite{pramanick2024spiqa}, M3Exam~\cite{zhang2023m3exam}, Math-Vision~\cite{wang2024measuring}, MATHCHECK-GEO~\cite{zhou2024your}, MathV360K~\cite{shi2024math}, MMSci~\cite{li2024mmsci}, DUDE~\cite{van2023document}, CMMMU~\cite{zhang2024cmmmu}, MathVista~\cite{lu2023mathvista}, ChartingNewTerritories~\cite{roberts2024charting}, CMMU~\cite{he2024cmmu}, MindBench~\cite{chen2024mindbench}, MathVerse~\cite{zhang2024mathverse}, NPHardEval4V~\cite{fan2024nphardeval4v}, SceMQA~\cite{liang2024scemqa}, CHOPINLLM~\cite{fan2024pre}, CharXiv~\cite{wang2024charxiv}, MMWorld~\cite{he2024mmworld}, MMTab~\cite{zheng2024multimodal}, VisualWebBench~\cite{liu2024visualwebbench}, ChartBench~\cite{xu2023chartbench}, MMC-Benchmark~\cite{liu2023mmc}, SciGraphQA~\cite{li2023scigraphqa} \\
\bottomrule
\end{tabular}}
\label{tab:sum_benchmark}
\end{table}

\section{Related Work}

\subsection{MLLM Benchmarks}

A variety of benchmark datasets have been proposed to evaluate the capabilities of MLLMs, covering visual perception, language understanding, external knowledge integration, and complex reasoning. To provide a more principled organization, we summarize representative multimodal benchmarks by their primary targeted capability, as shown in Table~\ref{tab:sum_benchmark}. Early datasets such as COCO~\cite{lin2014microsoft}, VQA~\cite{antol2015vqa}, and Visual7W~\cite{zhu2016visual7w} mainly assess perception and basic question answering abilities. Subsequent benchmarks like GQA~\cite{hudson2019gqa} and VCR~\cite{zellers2019recognition} emphasize compositional and commonsense reasoning, while OKVQA~\cite{marino2019ok} and TextVQA~\cite{singh2019towards} introduce challenges involving external knowledge and OCR-based understanding. ScienceQA~\cite{lu2022learn} further targets scientific reasoning in educational contexts.
As MLLMs~\cite{bai2025qwen2, zhu2025internvl3, hong2025glm} develop, these earlier benchmarks are hard capture their increasingly diverse capabilities. Recent efforts, such as MMBench~\cite{liu2024mmbench}, OmniBench~\cite{li2024omnibench}, and UniBench~\cite{al2024unibench}, aim to provide broader evaluations that include mathematics, medical diagnosis, chart understanding, and instruction following. However, these benchmarks typically rely on heuristic or empirical task categorization, leading to overlapping definitions, inconsistent classification, and limited diagnostic granularity, thus hindering efficiency analysis of model capabilities.

\subsection{Evaluation for Benchmarks}

In addition to evaluate the capabilities of MLLMs themselves, recent studies have begun to examine the design quality and diagnostic effectiveness of evaluation benchmarks. Most existing approaches follow a model-centric paradigm, where the quality of a benchmark is assessed indirectly through model performance across various tasks. Representative methods such as LIME~\cite{zhu2024lime} and CVR~\cite{zerroug2022benchmark} incorporate human annotations to interpret model errors and seek a balance between task difficulty and error susceptibility. However, these `Human Eval' approaches suffer from high labor costs and limited scalability. To address this limitation recent studies like Zerobench~\cite{roberts2025zerobench} and EnigmaEval~\cite{wang2025enigmaeval} have explored `model Eval' evaluation, which infers task difficulty and redundancy directly from model behavior. While more scalable, these methods risk introducing model-specific biases that compromise objectivity. Concurrently, efforts such as Information Density~\cite{li2025information} investigate properties like fallacy, difficulty, redundancy, and diversity from a data-centric perspective, and Redundancy Principles~\cite{zhang2025redundancy} offering new insights for benchmark design. Nonetheless, these approaches often rely on heuristic analyses and lack principled theoretical grounding. 

To bridge this gap, we propose a SEM-based framework that captures latent ability structures through causal modeling, enabling systematic evaluation of a benchmark’s construct validity and cognitive alignment. Building on this, we further introduce the \textsc{Gold} inspired by Piaget’s theory of cognitive development~\cite{piaget2000piaget}, organizing MLLM capabilities into a three-tier structure: Perception, Memory, and Reasoning, to enhance interpretability and theoretical coherence in benchmark design.

\section{Evaluating Benchmark Structures via SEM}
\label{sec:sem_framework}

\subsection{Revisiting of Structural Equation Modeling} 
\label{sec:revist}
SEM can be constructed based on causal measurement models~\cite{bollen2013eight}, organizing multiple observed variables into a coherent structural framework. It consists of two core components: the measurement model~\cite{neely1998three}, which defines the relationships between observed variables and their underlying latent constructs, and the structural model~\cite{lendaris2007structural}, which captures the causal relationships among latent variables.

\begin{figure}
  \centering
  \includegraphics[width=1\linewidth]{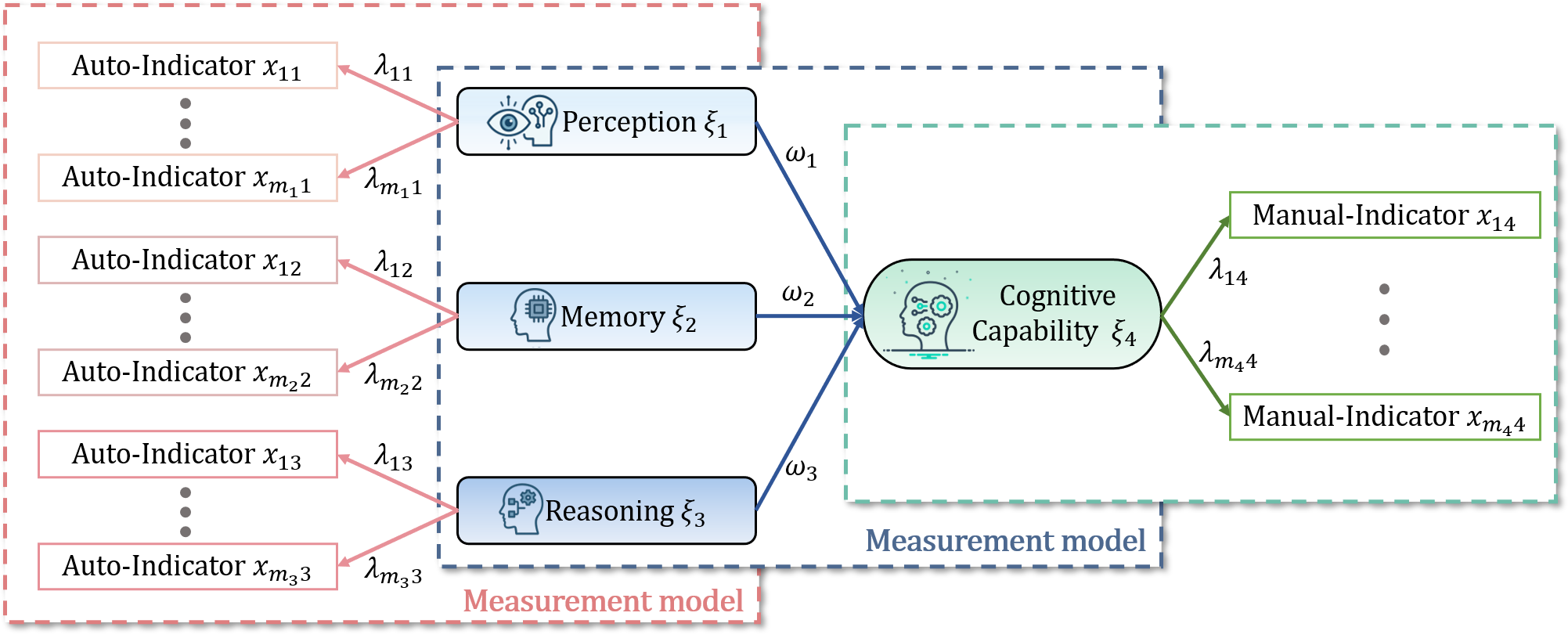}
  \caption{An architecture of SEM-based evaluation framework.}
  \label{fig:Gold_sem_system}
\end{figure}

For evaluating MLLMs capabilities, we focus on the model's performance across diverse tasks, where task scores collectively determine a latent ability. The specific ability does not preexist to drive task performance, it emerges from the aggregation of task-level outcomes. This paradigm aligns naturally with a formative measurement model, where multiple observed indicators jointly constitute the latent variable. Therefore, we adopt a formative measurement model for benchmark evaluation as shown in Fig.~\ref{fig:Gold_sem_system}. The evaluation framework defines the relationships between observed and latent variables as follows: $X = \Lambda_x \xi + \delta$ and $Y = \Lambda_y \eta + \epsilon$, where $X$ and $Y$ represent observed exogenous and endogenous variables, $\xi$ and $\eta$ denote latent variables, $\Lambda_x$ and $\Lambda_y$ are factor loading matrices, and $\delta$ and $\epsilon$ correspond to measurement errors. The structural model captures the causal relationships among latent variables: $\eta = B\eta + \Gamma \xi + \zeta$, where $B$ and $\Gamma$ are path coefficient matrices, and $\zeta$ represents residuals. Together, the measurement and structural models offer a comprehensive framework to assess both direct and indirect effects.

\begin{figure*}
  \centering
  \includegraphics[width=0.8\linewidth]{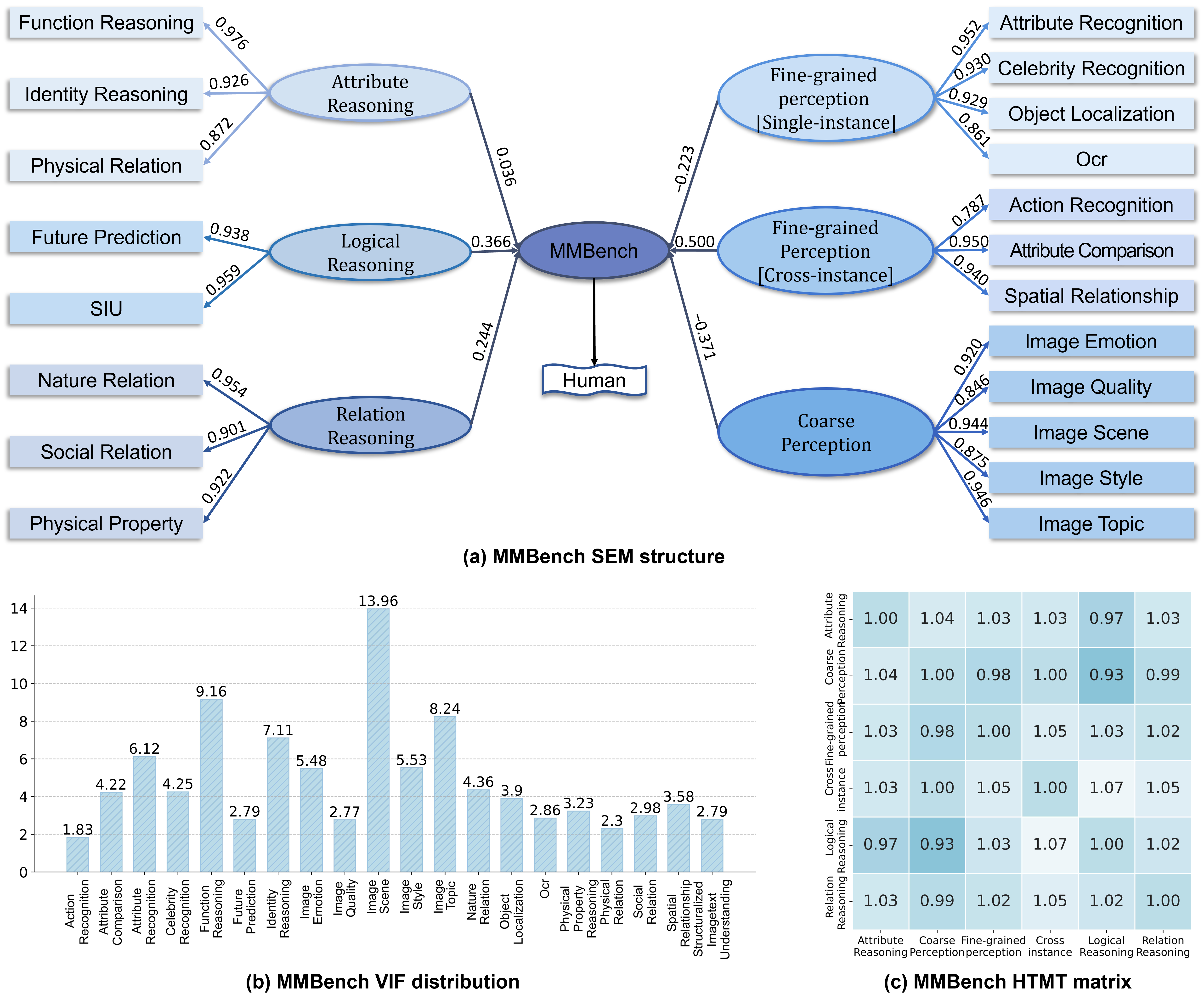}
  \caption{Structural analysis of MMBench by using SEM-based framework. MMBench shows stronger task contributions but suffers from redundancy.}
  \label{fig:MMBench_sem}
\end{figure*}

\subsection{Structural Analysis of Existing Benchmarks}

Interestingly, we further apply the proposed SEM-based evaluating framework to analyze existing MLLM benchmarks and find that most benchmarks struggle to balance low redundancy with high construct validity. Specifically, we investigate whether the task configurations within current benchmarks provide complementary information while jointly forming meaningful latent constructs. 

To illustrate this issue in detail, we conduct a case study on two representative benchmarks: MMBench~\cite{liu2024mmbench} and BLINK~\cite{fu2024blink}. Leveraging the performance reports of over 190 MLLMs collected via the VLMEvalKit~\cite{duan2024vlmevalkit} toolkit, we construct formative structural equation models based on the hierarchical task taxonomy specified by each benchmark. In our modeling framework, individual tasks serve as observed indicators, which are grouped into intermediate latent variables representing ability dimensions such as perception, reasoning, and cross-modal understanding. These first-order latent abilities are then aggregated into a second-order latent variable that captures the overall capability of an MLLM. Crucially, to align this latent construct with human-perceived performance, we incorporate the Chatbot Arena~\cite{chiang2024chatbot} score as an external observed indicator of the top-level latent variable. This hierarchical SEM structure enables us to evaluate how well benchmark-defined ability dimensions, as derived from task-level scores, explain human-judged overall performance.

In the case of MMBench, as shown in Fig.~\ref{fig:MMBench_sem}(a), tasks exhibit high contribution to their corresponding ability dimensions, with outer loadings exceeding 0.85 for most tasks. However, the analysis of Variance Inflation Factor (VIF) values in Fig.~\ref{fig:MMBench_sem}(b) shows a concerning level of multicollinearity, particularly for tasks like Image Scene (VIF = 13.96) and Attribute Recognition (VIF = 6.12), indicating a high degree of redundancy within the benchmark. This redundancy undermines the overall efficiency of the measurement model, as the tasks are provide insufficiently independent information.

\begin{figure*}
  \centering
  \includegraphics[width=0.85\linewidth]{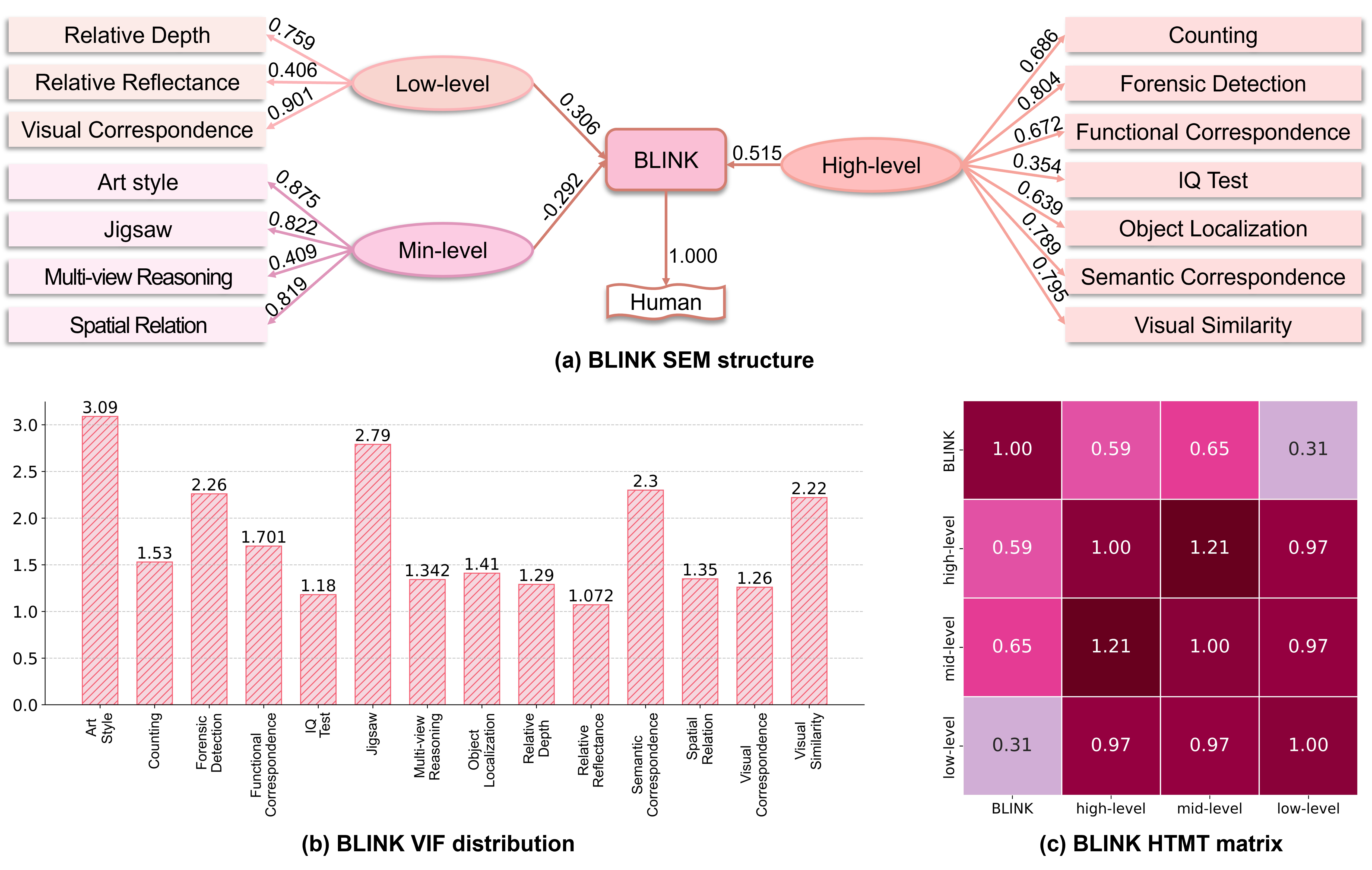}
  \caption{Structural analysis of BLINK by using SEM-based framework. BLINK has lower redundancy but weaker task contributions.}
  \label{fig:BLINK_sem}
\end{figure*}

On the other hand, BLINK as shown in Fig.~\ref{fig:BLINK_sem} exhibits a different issue. While the VIF values for BLINK are generally low (with a maximum value of 3.09 for Art Style), indicating low task redundancy, the outer loadings of several tasks are lower compared to MMBench. Tasks such as IQ Test (outer loading = 0.354) and Counting (outer loading = 0.686) contribute less to their respective constructs, suggesting that the benchmark lacks sufficiently strong task performance indicators for each cognitive ability dimension.

Despite structural differences, both MMBench and BLINK exhibit limited discriminant validity among their ability dimensions, as indicated by their respective HTMT matrices. 
Several HTMT values exceed or approach the 0.90 threshold, particularly between conceptually adjacent dimensions, revealing substantial overlap and suggesting that current ability taxonomies in both benchmarks need further conceptual refinement. For instance, in BLINK, the HTMT value between the mid-level and high-level reaches 1.21, highlighting that tasks intended to measure these distinct abilities are highly correlated and may not effectively disentangle the underlying cognitive constructs.

Overall, while MMBench and BLINK differ in their structural characteristics, MMBench showing strong but redundant task contributions, and BLINK presenting low redundancy but weaker indicators. Both suffer from poor discriminant validity in their ability dimensions. These findings expose fundamental limitations in current benchmark design, highlight the need for a more systematic approach to evaluating benchmark structure and the necessity for a principled framework to evaluate the internal structure of MLLM benchmarks and to redefine ability dimensions based on sound cognitive foundations.

\subsection{SEM-based Evaluation Framework for Benchmark}

To evaluate the robustness and interpretability of MLLM Benchmarks, we propose an evaluation framework for benchmark, which conducts a comprehensive concept-level analysis of the latent ability dimensions and their associated task indicators. It focuses on three key aspects: \textbf{dimensional diversity}, \textbf{task contribution} and \textbf{indicator validity}. These factors reflect the conceptual soundness and representational adequacy of the evaluation benchmark, as well as the redundancy and contribution of its indicators to the corresponding tasks.

\paragraph{Dimensional Diversity}
it aims to quantify the discriminant validity among latent ability dimensions, thereby characterizing their conceptual independence and potential redundancy. Although existing benchmarks effectively capture multiple facets of MLLMs’ capabilities, overlaps between different dimensions may still exist. To rigorously assess discriminant validity, we adopt the Heterotrait–Monotrait Ratio of correlations (HTMT)~\cite{dirgiatmo2023testing}. For any two latent dimensions $\xi_i$ and $\xi_j$, HTMT index is defined as
\begin{equation}
\text{HTMT}_{ij} = \frac{\frac{1}{|H_{ij}|} \sum\limits_{(x_p \in H_i, x_q \in H_j)} \text{cor}(x_p, x_q)}{\frac{1}{|M_{ii}|} \sum\limits_{(x_m, x_n \in H_i)} \text{cor}(x_m, x_n)},
\end{equation}
where $H_i$ and $H_j$ denote the sets of observed indicators corresponding to latent dimensions $\xi_i$ and $\xi_j$, respectively. The set $H_{ij}$ comprises all possible pairs $(x_p, x_q)$ such that $x_p \in H_i$ and $x_q \in H_j$, representing heterotrait--heteromethod comparisons. Each $x_p$ and $x_q$ denotes an observed variable (e.g., a task-specific score) that loads onto the latent dimensions $\xi_i$ and $\xi_j$, respectively. Conversely, the set $M_{ii}$ includes all within-construct pairs $(x_m, x_n)$ such that both $x_m$ and $x_n$ belong to $H_i$, representing monotrait--heteromethod comparisons. The function $\mathrm{cor}(\cdot, \cdot)$ denotes the Pearson correlation coefficient between observed variables.

Empirical guidelines suggest that an HTMT value exceeding 0.90 indicates deficient discriminant validity and substantial conceptual redundancy between constructs, whereas lower values imply satisfactory separation. Building on this heuristic, we define the dimensional diversity score as
\begin{equation}
D_{\mathrm{div}} =  \frac{1} {2 * \max_{i \neq j} \mathrm{HTMT}_{ij}},
\end{equation}
where $\max_{i \neq j} \mathrm{HTMT}{ij}$ represents the maximum HTMT value across all distinct pairs of latent dimensions. The denominator is scaled by 2 to ensure that $D_{\mathrm{div}} \in [0, 1]$, given that HTMT values are typically bounded within the empirical range of [0, 1]. A higher $D_{\mathrm{div}}$ indicates stronger discriminant validity and thus greater conceptual distinctiveness among latent constructs.

\paragraph{Task Contribution}  
it quantifies the relative importance of each task indicator in shaping its corresponding latent construct. In order to capture this notion in a formative measurement context, we adopt the concept of out‑loading~\cite{latan2013results}, which represents the standardized weight of each observed task on the latent dimension.
To evaluate the contribution of tasks at the level of the entire benchmark, which consists of multiple ability dimensions $\xi_{} = \{\xi_1, \dots, \xi_m\}$, we define the benchmark-level task contribution score as
\begin{equation}\label{eq:enhancement1}
\begin{split}
\mathrm{TC} & = \frac{1}{\sum_{i=1}^{n} m_i} \sum_{i=1}^{m_i} \sum_{j=1}^{n} \left| \lambda_{i,j} \right|,
\end{split}
\end{equation}
the metric $\mathrm{TC} \in [0,1]$ provides a global estimate of task-level informativeness across the entire benchmark, with higher values indicating stronger and more concentrated contributions from individual tasks.

\begin{table}[t]
\centering
\caption{Refined Task Selection Results on MMBench}
\resizebox{\columnwidth}{!}{
\begin{tabular}{lccc}
\toprule
    Capability & Representative Task & VIF & Outer Loading \\ \midrule
    Attribute  & Identity Reasoning & 1.749 & 0.893 \\
    Reasoning & Physical Relation & 1.749 & 0.924 \\ \midrule
    Coarse  & Image Quality & 2.682 & 0.938 \\
    Perception & Image Style & 2.682 & 0.954 \\ \midrule
    Cross-instance  & Action Recognition & 1.549 & 0.818 \\
    Perception & Spatial Relationship & 1.549 & 0.949 \\ \midrule
    Fine-grained  & Object Localization & 2.113 & 0.956 \\
    Perception & OCR & 2.113 & 0.896 \\ \midrule
    Logical  & Future Prediction & 2.793 & 0.938 \\
    Reasoning & Structured Image-Text Understanding & 2.793 & 0.959 \\ \midrule
    Relation  & Physical Property Reasoning & 2.133 & 0.941 \\
    Reasoning & Social Relation & 2.133 & 0.918 \\ \bottomrule
\end{tabular}
}
\label{tab:SEM_refined_MMBench}
\end{table}

\paragraph{Indicator Validity}
it is assessed both substantively and statistically, with the latter focusing on multicollinearity diagnostics among task indicators. For each indicator $x_{i,j}$ associated with latent dimension $\xi_j$, we compute the Variance Inflation Factor (VIF)~\cite{marcoulides2019evaluation} by
\begin{equation}
\mathrm{VIF}_{x_{i,j}} = \frac{1}{1 - R^2_{x_{i,j}}},
\end{equation}
where $R^2_{x_{i,j}}$ is the coefficient of determination obtained by regressing $x_{i,j}$ on the remaining tasks. A higher VIF indicates stronger collinearity, suggesting that task may not provide unique information beyond what is already captured by others.

To consolidate individual VIF values into a single benchmark-level measure, we define the \textit{composite validity score} \(D_{\mathrm{valid}}\) as the inverse of the geometric mean of all task-level VIFs:
\begin{equation}
D_{\mathrm{valid}}
= \left( \prod_{j=1}^{n} \mathrm{VIF}_{x_j} \right)^{-1/n},
\end{equation}
where \(n\) denotes the total number of task indicators in the benchmark. 
This aggregate score \(D_{\mathrm{valid}} \in (0,1]\) reflects the overall of indicator independence across the benchmark. Lower values indicate potential redundancy and multicollinearity issues at a global level, while higher values suggest stronger statistical validity of task indicators as formative constructs.

\section{\textsc{Gold} Benchmark}
\subsection{Motivation}
SEM analysis within existing benchmarks reveal that many tasks are characterized by \textbf{low contribution} to latent capability measurement and \textbf{high redundancy} with respect to other tasks. This observation raises a central concern: whether the increasing number of tasks truly enhances evaluation validity, or merely inflates benchmark scale without substantial benefit. A natural idea is to remove these redundant and low-contribution tasks. However, this immediately leads to a methodological challenge: would eliminating such tasks compromise the benchmark’s ability to capture and differentiate model capabilities? If so, task reduction would risk undermining benchmark reliability; if not, it could provide a principled way to improve evaluation efficiency. 

To address this issue, we first empirically verify that reducing the number of tasks does not harm the benchmark’s ability to represent model capabilities. However, we also observe that redundancy persists across capability dimensions, indicating that task-level pruning alone is insufficient. This highlights the necessity of introducing a theoretically grounded capability stratification. Therefore, we incorporate a hierarchical structure of capabilities inspired by Piaget’s cognitive development theory, which provides a principled foundation for disentangling and organizing evaluation dimensions.

\begin{table}[t]
\centering
\caption{Refined Task Selection Results on BLINK}
\resizebox{\columnwidth}{!}{
\begin{tabular}{lccc}
\toprule
        Capability & Representative Task & VIF & Outer Loading \\ \midrule
        Low-level & Relative Depth & 1.248 & 0.76 \\
         & Visual Correspondence & 1.248 & 0.921 \\ \midrule
         & Art Style & 2.884 & 0.869 \\
        Mid-level & Jigsaw & 2.730 & 0.816 \\
        & Spatial Relation & 1.318 & 0.829 \\ \midrule
       & Forensic Detection & 2.137 & 0.888 \\
        High-level & Semantic Correspondence & 1.962 & 0.881 \\ 
         & Visual Similarity & 2.015 & 0.852 \\ \bottomrule
    \end{tabular}}
\label{tab:SEM_refined_BLINK}
\end{table}

\subsection{Empirical Evidence}

\begin{table*}[h]
\centering
\caption{Reliability and Validity Metrics for Latent Constructs}
\label{tab:MMBench_BLINK_reliability}
\resizebox{1.9\columnwidth}{!}{
\begin{tabular}{lccccc}
\toprule
Benchmark & Construct & Cronbach's $\alpha$ & Reliability & Composite Reliability & Convergent Validity \\
\midrule
& Attribute Reasoning & 0.791 & 0.806 & 0.905 & 0.826 \\ 
& Coarse Perception & 0.884 & 0.898 & 0.945 & 0.895 \\ 
& Fine-grained perception & 0.841 & 0.945 & 0.924 & 0.858 \\ 
MMBench & Cross-instance Perception & 0.746 & 0.940 & 0.879 & 0.785 \\ 
& Logical Reasoning & 0.890 & 0.914 & 0.947 & 0.900 \\
& Relation Reasoning & 0.843 & 0.858 & 0.927 & 0.864 \\ \midrule
& high-level & 0.846 & 0.856 & 0.906 & 0.763 \\
BLINK & low-level & 0.717 & 0.726 & 0.831 & 0.713 \\
& mid-level & 0.800 & 0.838 & 0.876 & 0.703 \\
\bottomrule
\end{tabular}}
\end{table*}

\textbf{Experimental Setup.} 
We refine both MMBench~\cite{liu2024mmbench} and BLINK~\cite{fu2024blink} by removing tasks with $\mathrm{VIF} > 5$ or $\mathrm{outer loadings} < 0.8$~\cite{purwanto2021partial}. Here, outer loading refers to the correlation between an observed indicator (task) and its assigned latent construct, with higher values indicating that the task reliably reflects the underlying capability dimension. This ensures that only non-redundant and sufficiently representative tasks are retained for further SEM-based evaluation. In addition, we ensure that each latent capability dimension retains at least two tasks, 
which is necessary for computing VIF and outer loadings. If all tasks within a dimension fail to meet the VIF or outer loading thresholds, we retain the tasks that are relatively closest to meeting these criteria, so that each capability dimension remains represented in the SEM analysis. 

\textbf{Task Refinement Results.} 
Tables~\ref{tab:SEM_refined_MMBench} and~\ref{tab:SEM_refined_BLINK} summarize the retained tasks with their corresponding VIF and outer loadings. After refinement, all tasks achieve acceptable VIF values and high outer loadings. Especially, the VIF of the Physical Relation task decreases from $2.30$ to $1.75$ while its outer loading increases from $0.872$ to $0.924$, and similar improvements are observed across other tasks, achieving a better balance between multicollinearity and task representativeness. These results demonstrate that SEM filtering successfully removes redundant and low-contribution tasks while maintaining reliable indicators for latent capability measurement. 

\textbf{Reliability and Validity Results.}
Table~\ref{tab:MMBench_BLINK_reliability} reports the reliability and validity metrics for the latent constructs across MMBench and BLINK. Most constructs achieve acceptable thresholds for Cronbach’s $\alpha$ ($>0.70$)~\cite{tavakol2011making}, Composite Reliability ($>0.80$)~\cite{bacon1995composite}, and Convergent validity ($>0.70$)~\cite{carlson2012understanding}, confirming that the retained tasks provide consistent and reliable measurements of the intended capabilities.
Formally, Cronbach’s $\alpha$ is defined as:
\begin{equation}
\alpha = \frac{k}{k-1} \left( 1 - \frac{\sum_{i=1}^{k}\sigma_i^2}{\sigma_T^2} \right),
\end{equation}
where $k$ is the number of items, $\sigma_i^2$ is the variance of item $i$, and $\sigma_T^2$ is the variance of the total score. 

Composite reliability (CR) is calculated as:
\begin{equation}
\mathrm{CR} = \frac{(\sum_{i=1}^{k}\lambda_i)^2}{(\sum_{i=1}^{k}\lambda_i)^2 + \sum_{i=1}^{k}\theta_i},
\end{equation}
where $\lambda_i$ denotes the standardized factor loading of item $i$, and $\theta_i$ is its error variance.  

Convergent validity, typically assessed by the Average Variance Extracted (AVE), is given by:
\begin{equation}
\mathrm{AVE}_j=\frac{\sum_k\left[w_{p_i}^2\mathrm{var}\left(\xi_j\right)\right]}{\sum_k\left[w_{p_i}^2\mathrm{var}\left(\xi_j\right)\right]+\sum_k\left(1-w_{p_i}^2\right)},
\end{equation}
where $w_{p_i}$ denotes the standardized factor loading of the $i$-th indicator on latent construct $j$, and $\mathrm{var}(\xi_j)$ represents the variance of the latent construct.

For MMBench, constructs such as \textit{Logical Reasoning} ($\alpha = 0.890$, composite reliability $=0.947$, convergent validity $=0.900$) demonstrate strong internal consistency and convergent validity, indicating well-defined and coherent capability dimensions.
For BLINK, while \textit{high-level} capabilities exhibit good reliability and validity, the \textit{low-level} dimension shows relatively weaker values ($\alpha = 0.717$, convergent validity $=0.713$), suggesting room for further refinement. The refined benchmarks capture capability dimensions with satisfactory reliability and validity, making them suitable for systematic evaluation without redundancy.

\begin{figure}[t]
    \centering
    \includegraphics[width=0.85\linewidth]{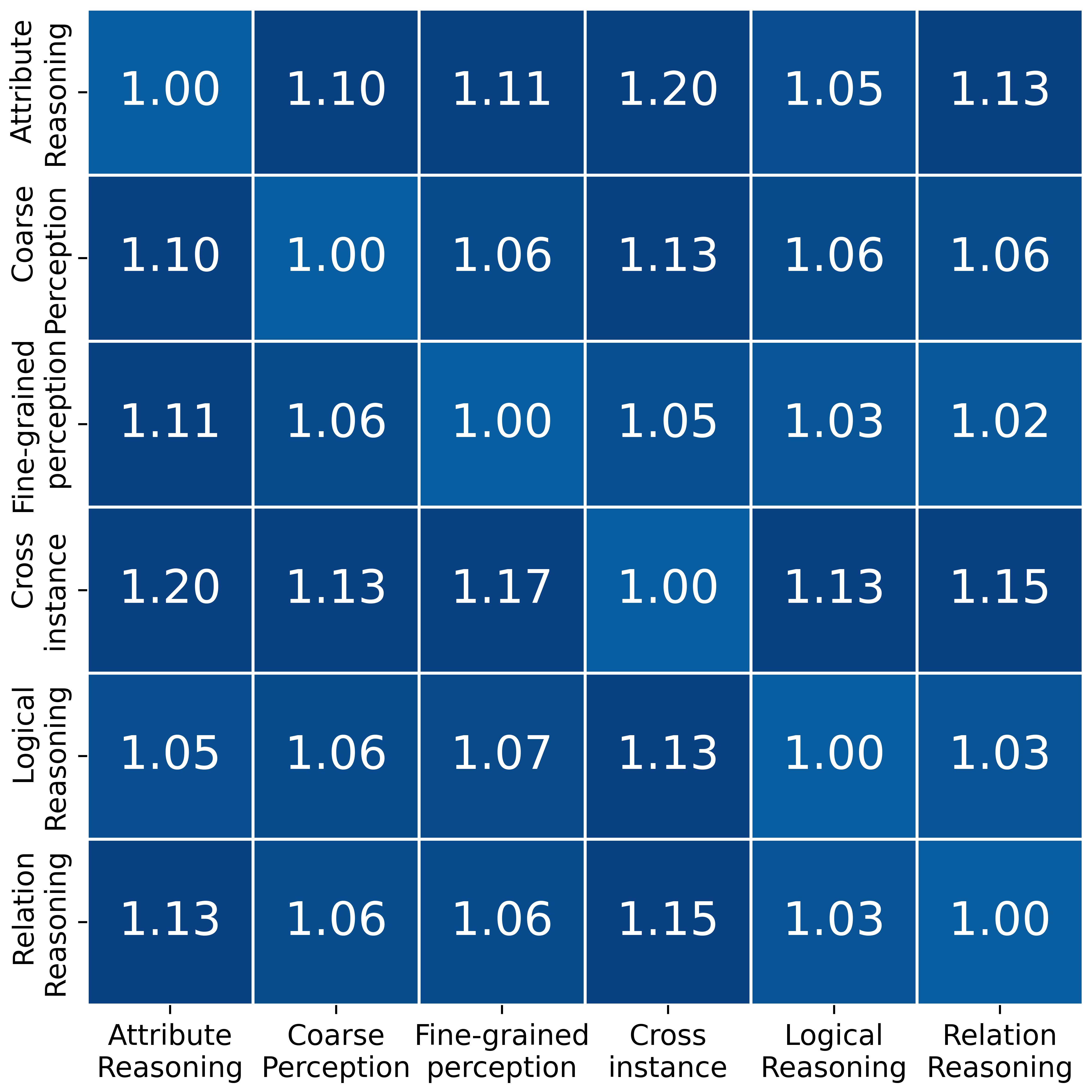}
    \caption{HTMT heatmap of MMBench latent capabilities.}
    \label{fig:HTMT_MMBench}
\end{figure}

\begin{table}[t]
\centering
\caption{HTMT Values of BLINK Latent Capabilities}
\resizebox{0.925\columnwidth}{!}{
\begin{tabular}{lcccc}
\toprule
& BLINK & High-level & Low-level & Mid-level \\
\midrule
BLINK & 1.00 & 0.50 & 0.68 & 0.36 \\
High-level & 0.50 & 1.00 & 1.12 & 0.98 \\
Low-level & 0.68 & 1.12 & 1.00 & 1.08 \\
Mid-level & 0.36 & 0.98 & 1.08 & 1.00 \\
\bottomrule
\end{tabular}}
\label{tab:HTMT_BLINK}
\end{table}

\begin{figure*}[t]
    \centering
    \includegraphics[width=\linewidth]{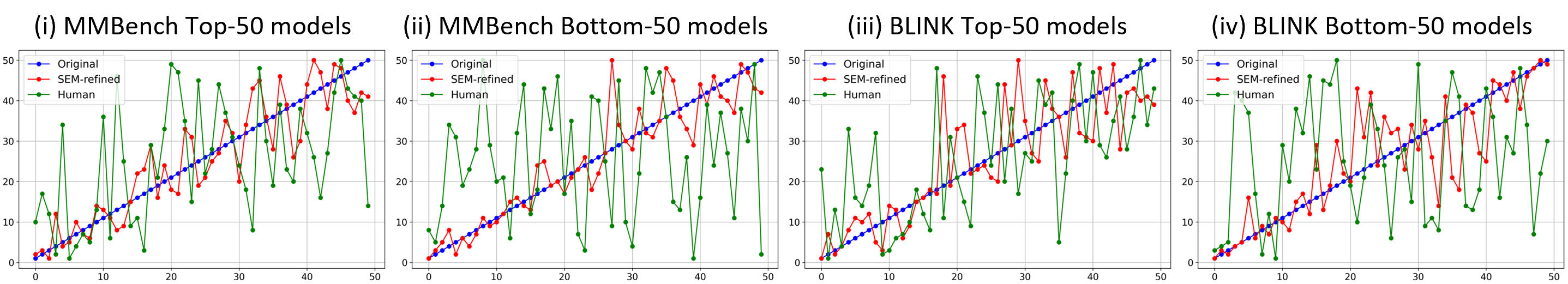}
    \caption{Rank correlation between original benchmarks (blue), SEM-refined benchmarks (red), and human evaluation (green). The four plots correspond to: (i) MMBench Top-50 models, (ii) MMBench Middle-50 models, (iii) BLINK Top-50 models, and (iv) BLINK Middle-50 models.}
    \label{fig:rank_corr}
\end{figure*}

\textbf{Discriminant Validity.} 
At the same time, an analysis of HTMT scores reveals that discriminant validity across latent capabilities remains limited. For MMBench as shown in Fig.~\ref{fig:HTMT_MMBench}, the cross-capability correlation between Cross-instance Perception and Attribute Reasoning increases from $1.03$ in the original benchmark to $1.20$ after SEM-based refinement, well above the recommended threshold of $0.90$. Similarly, as shown in Table~\ref{tab:HTMT_BLINK} BLINK retains high HTMT values, confirming that cross-level dependencies persist. This increase indicates that task-level pruning did not reduce inter-capability entanglement; rather, the latent capability dimensions have become even more overlapping. Such persistent and elevated correlations highlight the limitations of task-level refinement alone and underscore the need for a theoretically grounded capability hierarchy to improve discriminant validity.

\begin{table}[t]
\centering
\caption{Spearman Rank Correlations Between Original, SEM-refined of MMBench/BLINK and Human Evaluation}
\resizebox{\columnwidth}{!}{
\begin{tabular}{lcccc}
\toprule
Benchmark & Subset & Origin vs. SEM & Origin vs. Human & SEM vs. Human \\
\midrule
MMBench & Top-50    & 0.9099 & 0.4193 & 0.4975 \\
        & Bottom-50 & 0.9314 & 0.2766 & 0.3376 \\ \midrule
BLINK   & Top-50    & 0.8395 & 0.4335 & 0.6138 \\
        & Bottom-50 & 0.8508 & 0.1826 & 0.2401 \\
\bottomrule
\end{tabular}
}
\label{tab:spearman}
\end{table}

\textbf{Ranking Stability.} 
Ranking stability analysis is presented in Fig.~\ref{fig:rank_corr}, which illustrates the consistency of model rankings before and after task pruning across different benchmarks. As shown in Table~\ref{tab:spearman}, for MMBench, the correlation between original and SEM-refined benchmarks is $0.91$ for the top-50 models and $0.93$ for the bottom-50 models, indicating that task refinement preserves the overall ranking structure. The correlations with human evaluation increase from $0.42$ to $0.50$ for the top-50 models and from $0.28$ to $0.34$ for the bottom-50 models, showing improved alignment with human judgment. Similarly, for BLINK, the original vs. SEM-refined correlations are $0.84$ (top-50) and $0.85$ (bottom-50), while correlations with human evaluation rise from $0.43$ to $0.61$ for the top-50 models and from $0.18$ to $0.24$ for the bottom-50 models. Overall, these results demonstrate that SEM-based task filtering maintains ranking stability while increasing the benchmark’s consistency with human evaluation, validating both the effectiveness and reliability of the refined benchmark.

\subsection{Hierarchical Structure Based on Piaget's Developmental Theory}
\label{sec:piaget_framework}
Jean Piaget's theory of cognitive development is a seminal framework in developmental psychology, which describes how human cognitive abilities evolve through structured stages from infancy to adulthood. The classic stages include the Sensorimotor Stage (0--2 years), characterized by sensory and motor interactions with the environment; the Preoperational Stage (2--7 years), marked by symbolic thinking and basic memory development; the Concrete Operational Stage (7--11 years), in which logical reasoning about concrete events emerges; and the Formal Operational Stage (11+ years), defined by abstract and hypothetical reasoning. 

Recent studies~\cite{wang2024coglm, jeong2025scope} suggest that MLLMs exhibit cognitive capabilities roughly equivalent to the Concrete Operational stages of Piaget’s hierarchy. This indicates that MLLMs primarily demonstrate perceptual processing, memory encoding and retrieval, and elementary reasoning (e.g., object classification, spatial relations, and cause-effect inference), while higher-order abstract reasoning and metacognition associated with the Formal Operational Stage remain largely unattainable. Thus, Piaget’s framework provides a natural, theory-driven lens for organizing and evaluating MLLM capabilities, as it aligns with the current developmental scope of these models and offers interpretable cognitive layers for benchmarking.

Motivated by this observation, we propose a new capability hierarchy grounded in Piaget's theory of cognitive development. Unlike previous heuristic-based groupings, the proposed approach introduces a human-centered, theory-driven structure to model the cognitive underpinnings of MLLM performance. By anchoring the evaluation framework in developmental principles, we aim to achieve a more interpretable, hierarchical, and cognitively faithful representation of model abilities.

This Piaget-inspired cognitive structure is embedded into our benchmark design, with each task aligned to one of the three hierarchical cognitive layers (\textbf{Perception}, \textbf{Memory}, and \textbf{Reasoning}). To ensure that this theoretically grounded stratification is not only conceptually meaningful but also empirically valid, we employ SEM to assess its reliability and distinctiveness. SEM confirms the effectiveness of this organization, demonstrating improved dimensional separability, reduced task redundancy, and stronger measurement consistency compared to conventional heuristic groupings.

\subsection{\textsc{Gold} Construction}
To construct a human-aligned benchmark structure, we map tasks from existing multimodal datasets to cognitive processes defined in a Piaget-inspired framework (Section~\ref{sec:piaget_framework}). This process is formalized by using SEM, as outlined in Algorithm~\ref{alg:sem-benchmark}.
We begin by curating an indicator pool by collecting tasks from four widely adopted multimodal benchmarks: MME~\cite{fu2023mme}, MMMU-Bench~\cite{yue2024mmmu}, BLINK~\cite{fu2024blink} and MMBench~\cite{liu2024mmbench}. Each task is manually annotated and assigned to one cognitive layer based on its processing demands (\textit{e.g.,} OCR $\rightarrow$ Perception; landmark recall $\rightarrow$ Memory; domain‐specific inference $\rightarrow$ Reasoning).

We then apply PLS-SEM to estimate both the measurement and structural models in a unified, variance-based manner. Unlike covariance-based SEM, which emphasizes global model fit, PLS-SEM prioritizes maximizing explained variance of latent constructs and predictive relevance of the indicators. The algorithm proceeds iteratively through the following steps.

For each construct $\xi_j$, a latent score at iteration $t$ is obtained as a weighted composite of its observed indicators ${x_{ij}}{i=1}^{p_j}$:
\begin{equation}
\hat{\xi}_j^{(t)} = \sum_{i=1}^{p_j} \lambda_{ij}^{(t)} x_{ij}, \quad
\text{with } \sum_{i=1}^{p_j} (\lambda_{ij}^{(t)})^2 = 1.
\end{equation}
Here, initial weights are set uniformly, and are subsequently refined based on inner and outer approximations.

The inner approximation aggregates information from adjacent constructs in the structural model. Specifically, a proxy $z_j^{(t)}$ is constructed as
\begin{equation}
z_j^{(t)} = \sum_{k \in \mathrm{Neigh}(j)} \alpha_{jk}^{(t)} \hat{\xi}k^{(t)},
\end{equation}
where $\alpha{jk}^{(t)} = \mathrm{sgn}\big(\mathrm{corr}(\hat{\xi}_j^{(t)}, \hat{\xi}_k^{(t)})\big)$ preserves the empirical directionality of associations.

The outer approximation updates the weights through an Ordinary Least Squares regression of indicators on the proxy score:
\begin{equation}
\lambda_{ij}^{(t+1)} = \frac{\mathrm{cov}(x_{ij}, z_j^{(t)})}{\sqrt{\mathrm{var}(x_{ij}) , \mathrm{var}(z_j^{(t)})}}.
\end{equation}
The weights are then normalized to unit length for numerical stability:
\begin{equation}
\lambda_{ij}^{(t+1)} \leftarrow \frac{\lambda_{ij}^{(t+1)}}{\sqrt{\sum_{k=1}^{p_j} (\lambda_{kj}^{(t+1)})^2}}.
\end{equation}

The procedure alternates between latent score estimation and weight updating until convergence, defined as
\begin{equation}
\max_{i,j} , |w_{ij}^{(t+1)} - w_{ij}^{(t)}| < \epsilon,
\end{equation}
or equivalently, stability in both indicator weights and structural path coefficients.

Upon convergence, the outer loadings
\begin{equation}
\lambda_{ij} = \frac{\mathrm{cov}(x_{ij}, \hat{\xi}_j)}{\sqrt{\mathrm{var}(x_{ij}) , \mathrm{var}(\hat{\eta}_j)}}
\end{equation}
quantify the reliability of each indicator, while structural path coefficients are obtained by regressing each endogenous construct on its predictors via OLS. These coefficients capture the hypothesized causal relations among latent capabilities, whereas the loadings inform indicator validity and redundancy.

\begin{algorithm}
\caption{Cognitive Benchmark Construction via Piaget Theory and SEM}
\label{alg:sem-benchmark}
\KwIn{Normalized task indicators $\mathcal{X}$ from various benchmarks}
\KwData{VIF threshold $\delta_{\text{VIF}}$, outer loading threshold $\lambda_{\min}$}
\BlankLine
\emph{// Stage 1: Initialization} \\
Define latent constructs: Perception, Memory, Reasoning\; 
Assign each task $x_i \in \mathcal{X}$ to its latent construct\; 
Initialize weights $w \gets 1$\;
\BlankLine
\emph{// Stage 2: Iterative Refinement} \\
\While{task with $\mathrm{VIF} > \delta_{\text{VIF}}$ or $\mathrm{outer\ loading} < \lambda_{\min}$}{
    1. Estimate measurement model via PLS-SEM (maximize explained variance)\;
    2. Remove redundant tasks with $\mathrm{VIF} > \delta_{\text{VIF}}$\;
    3. Remove low-quality tasks with $\mathrm{outer\ loading} < \lambda_{\min}$\;
}
\BlankLine
\emph{// Stage 3: Validation} \\
1. Validate model reliability using Cronbach’s $\alpha$\;
2. Assess discriminant validity using HTMT\;
\KwOut{Theory-aligned benchmark structure with validated task indicators}
\end{algorithm}

\begin{figure*}
  \centering
  \includegraphics[width=0.8\linewidth]{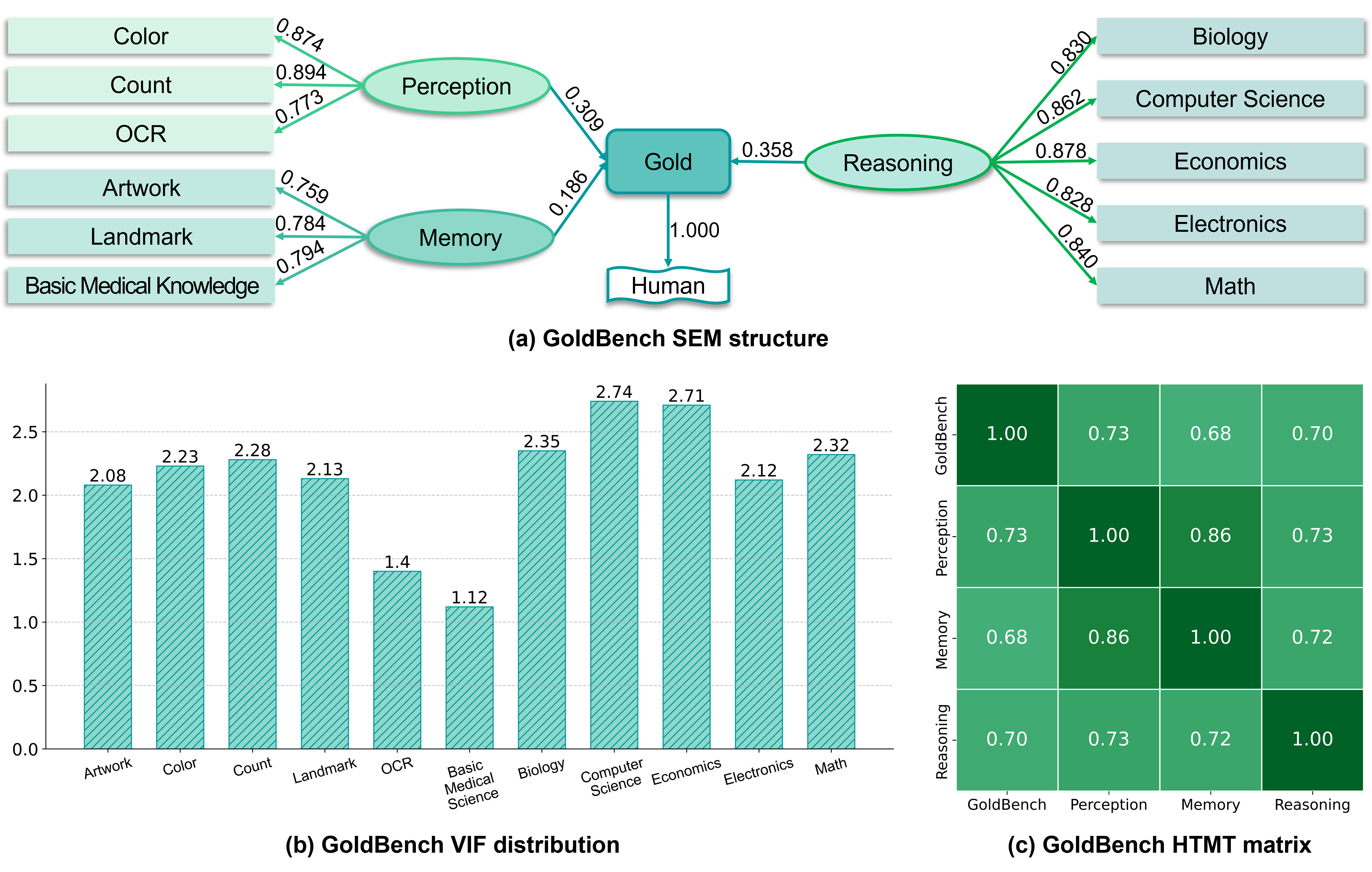}
  \caption{Structural analysis of \textsc{Gold} by using SEM-based framework.}
  \label{fig:Gold_sem}
\end{figure*}

\begin{table}[h]
\centering
\caption{Reliability and Validity Metrics for Latent Constructs}
\label{tab:sem_reliability}
\resizebox{1\columnwidth}{!}{
\begin{tabular}{l|ccc}
\toprule
    \multirow{2}{*}{\makecell{Metrics}} & \multicolumn{3}{c}{\makecell{Latent Constructs}} \\
     & Memory & Perception & Reasoning \\ \midrule
    Cronbach's $\alpha$ & 0.702 & 0.804 & 0.902 \\
    Reliability & 0.732 & 0.813 & 0.906 \\
    Composite reliability & 0.823 & 0.885 & 0.927 \\
    Convergent validity & 0.707 & 0.720 & 0.719 \\ \midrule
    SRMR & ~ & 0.087 & ~ \\ \midrule
    $R^2$ & ~ & 0.557 & ~ \\
\bottomrule
\end{tabular}}
\end{table}

To refine the model, we iteratively prune the indicator pool by removing tasks exhibiting high multicollinearity (variance inflation factor $> \delta_{\text{VIF}}$) or insufficient indicator quality (outer loading $< \lambda_{\min}$). Following prior empirical standards~\cite{fong2013hair, 10990127}, we set $\delta_{\text{VIF}} = 5$ and $\lambda_{\min} = 0.75$. This process enhances both statistical validity and conceptual coherence. We further evaluate construct reliability and discriminant validity using Cronbach’s $\alpha$ and the heterotrait–monotrait (HTMT) ratio.

The final output, \textsc{Gold}, is a compact and theory-grounded task set with optimized indicator weights and path coefficients. Each selected task robustly represents its assigned cognitive function, enabling interpretable, scalable, and semantically aligned evaluation of MLLMs.

\begin{table*}[t]\setlength\tabcolsep{3pt}
    \centering
    \caption{Comparison of State-of-the-Art Methods on \textsc{Gold}}
    \resizebox{0.95\textwidth}{!}{
		\begin{tabular}{l r r ccc ccc ccccc}
			\toprule
			\multirow{2}{*}{Methods} & \multirow{2}{*}{\makecell{Model \\ size}} &\multirow{2}{*}{Overall} & \multicolumn{3}{c}{Perception} & \multicolumn{3}{c}{Memory} & \multicolumn{5}{c}{Reasoning}  \\ 
             \cmidrule(r){4-6} \cmidrule(r){7-9} \cmidrule(r){10-14} 
			& & & Color  & Count & OCR & Artwork & Landmark & BMK & Biology & CS & Economics & Electronics & Math   \\
			\midrule
			\multicolumn{14}{l}{\textbf{MLLM (closed source)}} \\ 
			\midrule
			\includegraphics[height=0.9em]{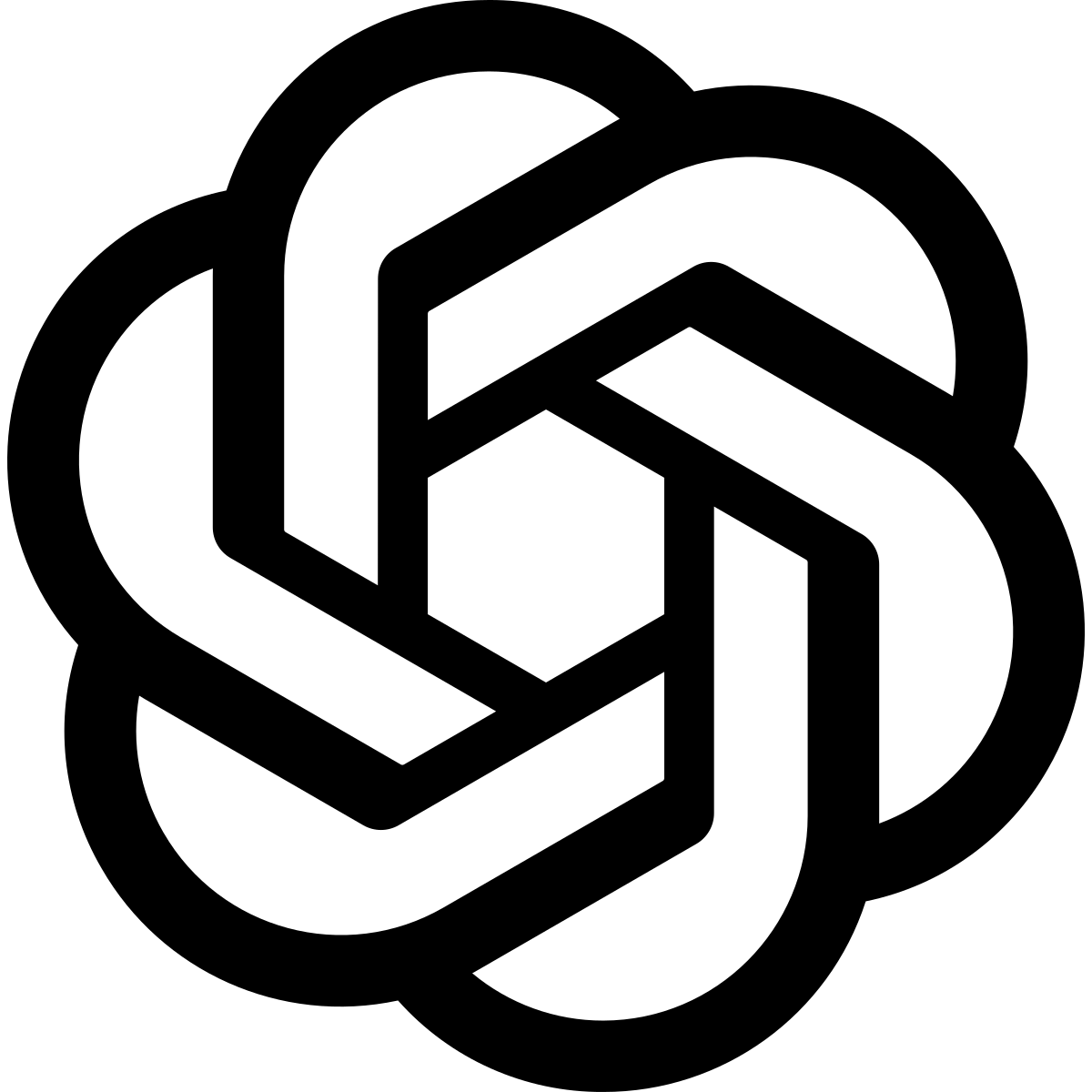} ~GPT-4o & - &216.3 & 92.5 & \textbf{92.5} & \textbf{96.3} & 72.6 & 91.0 & 73.3 & 56.7 & 73.3 &80.0 & 76.7 & \textbf{70.0} \\
            \includegraphics[height=0.9em]{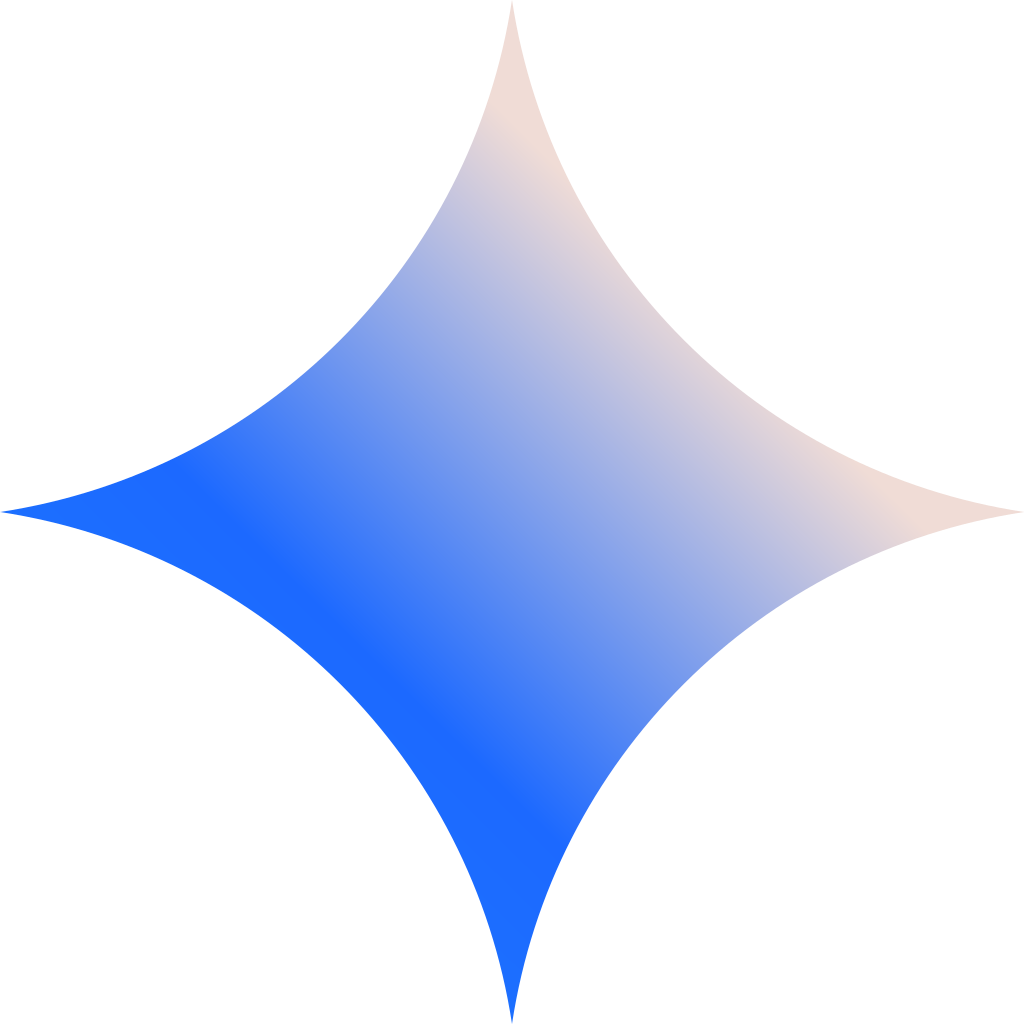} ~Gemini-2.0-Pro & - & 212.9  & 90.0  & 82.5  & 88.8  & 81.0  & 92.8  & \textbf{83.3}  & 60.0  & 70.0  & \textbf{83.3}  & 83.3 & 56.7  \\
            \includegraphics[height=0.9em]{./logo/gemini-color.png} ~Gemini-2.0-Flash & - & 215.5  & \textbf{95.0}  & 82.5  & \textbf{96.3}  & 73.5  & 84.9  & 60.0  & 60.0  & \textbf{83.3}  & 80.0  & \textbf{86.7}  & 60.0  \\
            \includegraphics[height=0.9em]{./logo/gemini-color.png} ~Gemini1.5-Pro & - & 183.9  & 90.0  & 77.5  & 68.8  & 72.8  & 74.6  & 63.3  & 46.7  & 66.7  & 70.0  & 53.3  & 63.3 \\ 
			\midrule
			\multicolumn{13}{l}{\textbf{Open source}} \\ 
			\midrule
                \includegraphics[height=0.9em]{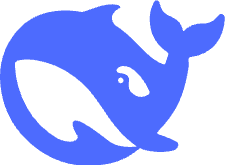}~DeepSeek-VL2-tiny & 3B & 156.2  & 66.7  & 71.7  & 81.3  & 63.5  & 84.3  & 40.0  & 40.0  & 40.0  & 53.3  & 53.3  & 50.0  \\
                \includegraphics[height=0.9em]{./logo/deepseek.png}~DeepSeek-VL2-small & 16B & 165.8  & 82.5  & 70.9  & 85.0  & 69.0  & 84.9  & 53.3  & 53.3  & 53.3  & 36.7  & 46.7  & 53.3  \\
                \includegraphics[height=0.9em]{./logo/deepseek.png}~DeepSeek-VL2 & 27B & 173.8  & 90.0  & 70.9  & 73.8  & 64.9  & 86.1  & 66.7  & 60.0  & 60.0  & 36.7  & 53.3  & 56.7  \\
                \includegraphics[height=0.9em]{./logo/deepseek.png}~DeepSeek & 1B & 118.2  & 62.5  & 56.7  & 40.0  & 62.3  & 78.8  & 46.7  & 30.7  & 46.7  & 36.7  & 30.0  & 16.7 \\
                \includegraphics[height=0.9em]{./logo/deepseek.png}~DeepSeek & 7B & 140.7  & 85.0  & 79.2  & 47.5  & 58.6  & 82.3  & 56.7  & 27.3  & 43.3  & 36.7  & 43.3  & 33.3 \\
                \includegraphics[height=0.9em]{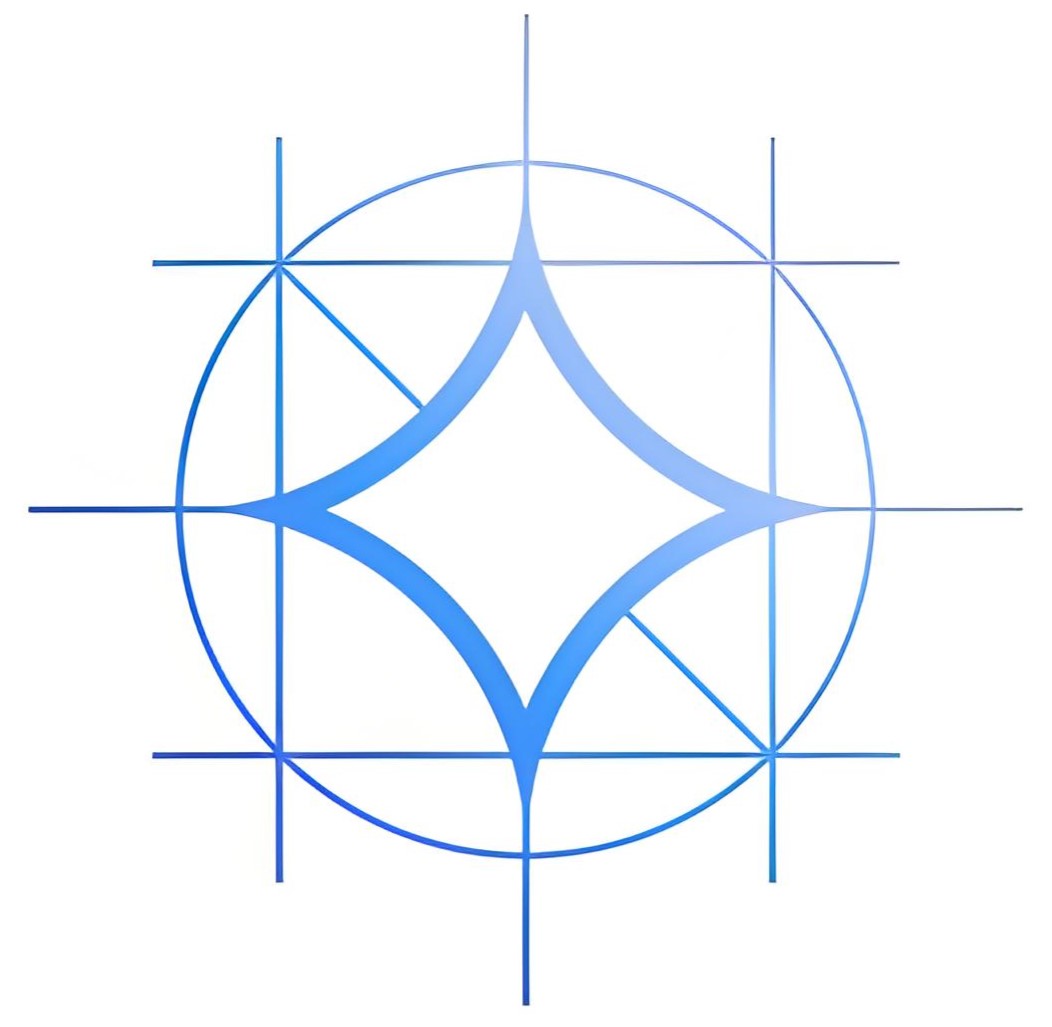} ~Gemma3 & 4B & 148.4  & 54.2  & 55.0  & 81.3  & 62.5  & 80.1  & 53.3  & 36.7  & 46.7  & 46.7  & 53.3  & 50.0  \\
                \includegraphics[height=0.9em]{./logo/gemma3.jpg} ~Gemma3 & 12B & 180.4  & 75.0  & 75.0  & 85.0  & 75.6  & 88.3  & 63.3  & 33.3  & 60.0  & 73.3  & 63.3  & 53.3  \\
                \includegraphics[height=0.9em]{./logo/gemma3.jpg} ~Gemma3 & 27B & 194.2  & 73.4  & 72.5  & 92.5  & 64.4  & 86.5  & 70.0  & 40.0  & 70.0  & 70.0  & \textbf{86.7} & 63.3 \\
                \includegraphics[height=0.9em]{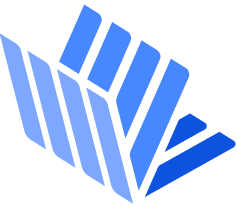} ~InternVL3 & 2B & 161.8  & 86.7  & 77.5  & 85.0  & 78.1  & 84.5  & 30.0  & 40.0  & 40.0  & 56.7  & 50.0  & 40.0  \\
                \includegraphics[height=0.9em]{./logo/InternVL.png} ~InternVL3 & 9B & 201.6  & 89.2  & 86.7  & \textbf{96.3}  & 80.9  & 87.9  & 63.3  & \textbf{66.7}  & 63.3  & 60.0  & 73.3  & 56.7  \\
                \includegraphics[height=0.9em]{./logo/InternVL.png} ~InternVL3 & 38B & 209.9  & 86.7  & 82.5  & \textbf{96.3}  & \textbf{82.4}  & 92.3  & 76.7  & 50.0  & 70.0  & 76.7  & 83.3  & 63.3  \\
                \includegraphics[height=0.9em]{./logo/InternVL.png} ~InternVL3 & 78B & \textbf{221.1}  & 91.7  & 85.0  & \textbf{96.3}  & 82.1  & \textbf{92.9}  & \textbf{83.3}  & \textbf{66.7}  & 80.0  & 80.0  & 76.7  & 66.7  \\
                \includegraphics[height=0.9em]{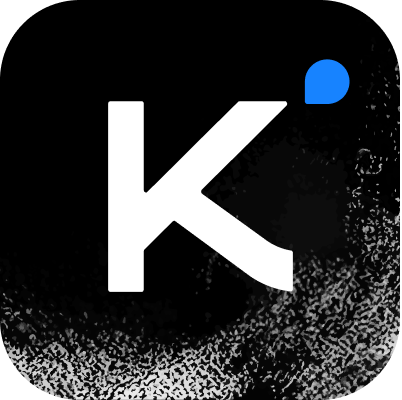} ~KimiVL & 16B & 183.9  & 92.5  & 85.0  & 77.5  & 72.8  & 89.5  & 53.3  & 46.7  & 63.3  & 73.3  & 63.3  & 53.3 \\
                \includegraphics[height=0.9em]{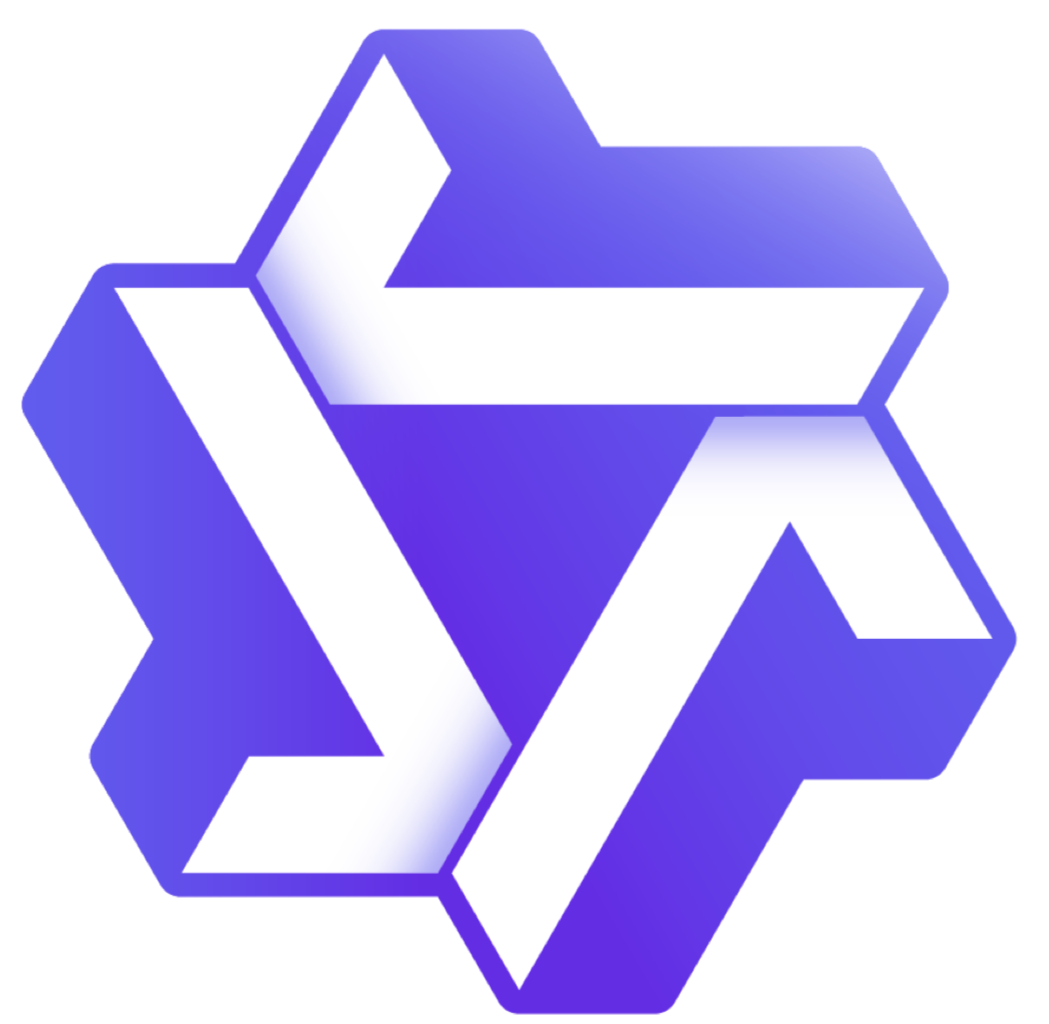} ~Qwen2.5-VL & 3B & 160.1  & 85.0  & 85.0  & 66.3  & 72.3  & 90.1  & 66.7  & 43.3  & 50.0  & 40.0  & 53.3  & 26.7  \\
                \includegraphics[height=0.9em]{./logo/qwen.png} ~Qwen2.5-VL & 7B & 195.9  & 92.5  & 82.5  & \textbf{96.3}  & 73.8  & 91.6  & 63.3  & 43.3  & 56.7  & 63.3  & 76.7  & 63.3  \\
                \includegraphics[height=0.9em]{./logo/qwen.png} ~Qwen2.5-VL & 72B & 207.2  & \textbf{95.0}  & 86.7  & \textbf{96.3}  & 70.4  & 79.8  & 73.3  & 60.0  & 63.3  & 80.0  & 83.3  & 50.0 \\
			\bottomrule
		\end{tabular}
	}
	\label{tab:sota_method}
\end{table*}

\subsection{Statistics.}
The final \textsc{Gold} Benchmark contains 12 tasks, selected from an initial pool of 85 across the four benchmarks. Our analysis follows the guidelines proposed by J. Hair~\cite{hair2019use}. For measurement model, we examine the indicator weights to evaluate their contribution to the latent constructs. As depicted in the Fig.~\ref{fig:Gold_sem}, the weights are consistently greater than 0.75. This outcome indicates that the retained indicators possess strong explanatory power and are effectively capturing the underlying constructs they are intended to measure.

We also assess multicollinearity using VIF. As shown in Fig.~\ref{fig:Gold_sem}(b), all indicators yield VIF values well below the conservative threshold of 5 (ranging from 1.12 to 2.75), suggesting the absence of problematic multicollinearity and supporting the discriminant structure of the model.
Finally, we evaluate discriminant validity using the Heterotrait-Monotrait (HTMT) ratio in Fig.~\ref{fig:Gold_sem}(c). All pairwise HTMT values are below the recommended cutoff of 0.90, with the highest observed value being 0.86 (between Memory and Perception). This result confirms that each latent construct captures a distinct cognitive domain and is not confounded with others.

To further verify the statistical quality of the benchmark, we conduct standard reliability and validity tests on the latent constructs. As shown in Table~\ref{tab:sem_reliability}, all constructs meet the recommended thresholds for internal consistency and convergent validity. Specifically, Cronbach’s $\alpha$ values are all above 0.70, reflecting acceptable internal reliability. Composite reliability scores exceed 0.80 across all dimensions, confirming the coherence of the grouped indicators. In terms of convergent validity, all constructs exhibit Average Variance Extracted (AVE) values above 0.60, suggesting that each construct explains a substantial portion of the variance in its associated indicators.

In addition to reliability and validity metrics, we also evaluate the standardized root mean square residual (SRMR) and the $R^2$ value for each latent construct, as presented in Table~\ref{tab:sem_reliability}. SRMR provides an absolute measure of model fit, indicating how well the observed data align with the theoretical model. For the Perception construct, SRMR is 0.087, slightly exceeding the conventional threshold of 0.08, suggesting an overall good fit. The $R^2$ value reflects the explanatory power of the model, with Perception achieving $R^2 = 0.557$, indicating that more than half of the variance in its indicators is accounted for by the latent construct. Memory and Reasoning also exhibit strong explanatory power, with Cronbach's $\alpha$, composite reliability, and convergent validity all exceeding recommended thresholds. Together, these metrics confirm that the \textsc{Gold} benchmark possesses robust reliability, valid measurement of latent constructs, and sufficient explanatory power to capture the underlying cognitive dimensions.

\section{Experimental Study}
\label{sec:results}

\subsection{Experiment Setting}

To evaluate the quality and interpretability of the \textsc{Gold} benchmark, we design a series of experiments involving both open-source and closed-source MLLMs. Our evaluation includes two core settings: (1) \textbf{Main Experiment}, where we assess mainstream models using a standardized pipeline to reveal overall performance differences; (2) \textbf{SEM-based Benchmark Evaluation Experiment}, where we validate \textsc{Gold}'s diagnostic effectiveness by measuring key metrics such as dimensional diversity ($D_\mathrm{div}$), task contribution ($\mathrm{TC}$), and indicator validity ($D_\mathrm{valid}$). We further assess its interpretability by comparing these outputs with human ratings to examine alignment with human judgment.

\textbf{Evaluation Models.} 
For main experiment, we evaluate both closed-source and open-source MLLMs. For the closed-source group, we include \textbf{GPT-4o}~\cite{achiam2023gpt}, \textbf{Gemini-2.0}~\cite{team2025gemini} and~\textbf{Gemini-1.5}~\cite{team2024gemini}. For the open-source group, we consider a range of models with varying scales, including \textbf{DeepSeekVL2}~\cite{wu2024deepseek}, \textbf{DeepSeek}~\cite{liu2024deepseek},  \textbf{Gemma3}~\cite{team2025gemma}, \textbf{InternVL3}~\cite{zhu2025internvl3}, \textbf{KimiVL}~\cite{team2025kimi} and \textbf{Qwen2.5-VL}~\cite{bai2025qwen2}. These models represent the current frontier of multimodal LLM development and cover diverse architecture families and parameter sizes.
For the benchmark evaluation experiment, we leverage evaluation results from 165 multimodal language models collected via VLMEvalKit~\cite{duan2024vlmevalkit}, which provides standardized testing across multiple benchmarks. The Metrics used in the experiment will be described in detail in the Appendixes.

\subsection{Main Results}

In this section, we report the evaluation outcomes of various open-source and closed-source MLLMs on the \textsc{Gold} benchmark under the \textit{Main Experiment} setting. Table~\ref{tab:sota_method} presents the detailed results, where models are assessed across three cognitive layers—\textbf{Perception}, \textbf{Memory}, and \textbf{Reasoning}—further subdivided into interpretable subtasks such as color recognition, counting, OCR, basic medical knowledge (BMK), and domain-specific logical reasoning. BMK and CS denote Basic Medical Knowledge and Computer Science, respectively. Overall scores are computed by weighting capability-level scores according to task contributions.

\textbf{Overall Performance.} 
Among closed-source models, GPT-4o achieves the highest overall score of 216.3, excelling particularly in Perception (Color 92.5, Count 92.5, OCR 96.3) and Memory tasks (Artwork 72.6, Landmark 91.0, BMK 73.3). Gemini-2.0 variants (Pro: 212.9, Flash: 215.5) also perform strongly, with Gemini-2.0-Flash achieving the highest OCR score (96.3) among closed-source models, while Gemini-2.0-Pro leads in Memory (Landmark 92.8, BMK 83.3). Among open-source models, InternVL3-78B reaches 221.1 overall, surpassing all other models including GPT-4o, while Qwen2.5-VL-72B achieves 207.2 overall, showing substantial improvements over smaller variants (3B: 160.1, 7B: 195.9). 

\textbf{Capability-Level Comparison.} 
Perception tasks are consistently the strongest across all models. For example, InternVL3-78B scores 91.7 (Color), 85.0 (Count), and 96.3 (OCR). DeepSeek models, though smaller (1B–27B), also show solid Perception performance, with DeepSeek-VL2 achieving 90.0 (Color) and 86.1 (Landmark), demonstrating that even lightweight open-source models can handle low-level visual understanding. \textbf{Memory} tasks show more variability: InternVL3-78B achieves 92.9 on Landmark but only 83.3 on BMK, while Qwen2.5-VL-72B scores 79.8 on Landmark and 73.3 on BMK. DeepSeek-VL2-small (16B) reaches 84.9 on Artwork and 53.3 on BMK, highlighting a gap between small-to-medium and large models. \textbf{Reasoning} tasks remain the most challenging across all families, with GPT-4o scoring 70.0 in Math and 80.0 in Economics, InternVL3-78B scoring 66.7 and 80.0 respectively, and DeepSeek models struggling further, e.g., DeepSeek 1B scoring only 16.7 in Math.

\textbf{Model Size and Performance.} 
Performance generally scales with model size. InternVL3 demonstrates steady gains from 161.8 (2B) to 221.1 (78B). Qwen2.5-VL improves from 160.1 (3B) to 207.2 (72B), while DeepSeek scales from 118.2 (1B) to 173.8 (27B), showing that even smaller architectures benefit from parameter growth. However, Reasoning subfields such as Electronics, Math, and Economics remain bottlenecks for all families.

\subsection{SEM-based Benchmark Evaluation Result Analysis}


\begin{figure}[t]
    \centering
    \includegraphics[width=1\linewidth]{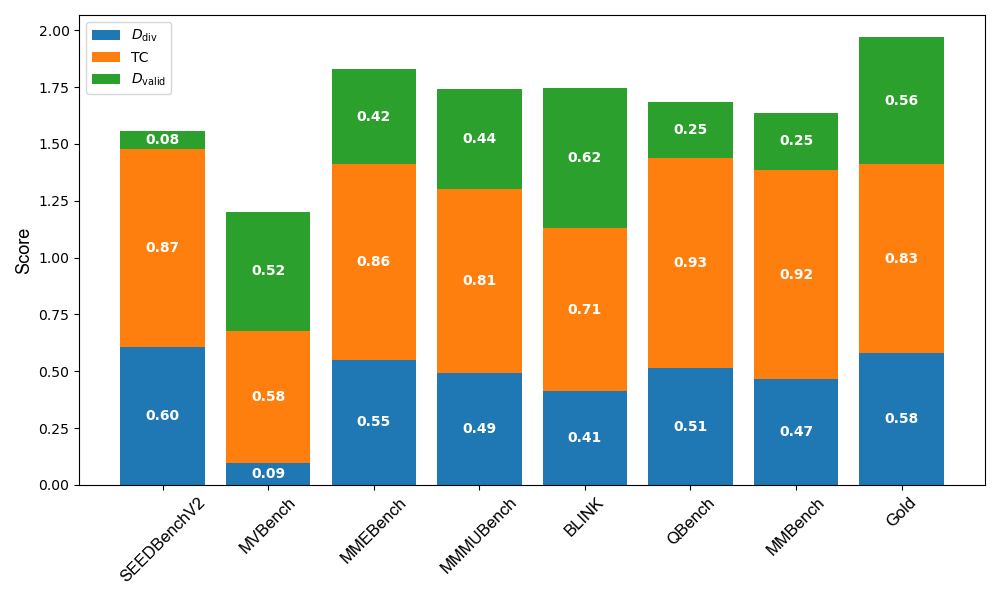}
    \caption{Diagnostic metrics for existing MLLM benchmarks across four dimensions.}
    \label{fig:Diagnostic metrics}
\end{figure}

As shown in Fig.~\ref{fig:Diagnostic metrics}, \textsc{Gold} achieves the best balance among the three evaluation dimensions, with the highest overall score (2.625). It combines broad cognitive coverage, strong task informativeness, and robust statistical validity, demonstrating the strength of theory-driven benchmark restructuring under the SEM-based framework. In contrast, some existing benchmarks show localized strengths but lack holistic consistency. Moreover, \textsc{Gold} shows the highest alignment with human ratings (Pearson $r = 0.7359$), reflecting its superior interpretability and evaluation reliability.

\textbf{Dimensional Diversity.}
$D_{\mathrm{div}}$ measures distinct and non-overlapping the latent ability dimensions are. It is computed as the inverse of the highest Heterotrait–Monotrait (HTMT) ratio across all construct pairs. Benchmarks like {SEEDBenchV2} (0.605), {MMEBench} (0.549), and our \textsc{Gold} (0.579) exhibit strong discriminant validity across dimensions, reflecting diverse and well-separated cognitive constructs. In contrast, {MVBench} (0.094) shows the lowest $D_{\mathrm{div}}$, suggesting highly entangled constructs that may compromise diagnostic clarity.

\textbf{Task Contribution.}
$\mathrm{TC}$ assesses the average magnitude of standardized loadings for all indicators across their respective constructs. Higher $\mathrm{TC}$ values suggest that tasks substantially shape latent abilities while minimizing redundancy. Most benchmarks perform well on this metric, including {QBench} (0.926), {MMBench} (0.920), and {MMEBench} (0.861). Our \textsc{Gold} benchmark (0.833) maintains a high $\mathrm{TC}$ while avoiding excessive task overlap. In contrast, {MVBench} again underperforms (0.584), indicating weaker task-level informativeness and possible construct inflation due to redundant indicators.

\textbf{Indicator Validity.}
$D_{\mathrm{valid}}$ evaluates how independent task indicators are in terms of multicollinearity. Computed as the inverse geometric mean of their VIFs, higher scores imply stronger statistical uniqueness across tasks. {BLINK} (0.618), \textsc{Gold} (0.557), and {MVBench} (0.522) exhibit the highest indicator validity, suggesting that their tasks are statistically independent and less redundant. Meanwhile, {SEEDBenchV2} (0.081) and {QBench} (0.246) show weak indicator validity, indicating significant multicollinearity among tasks that may hinder interpretability.

\begin{figure}
    \centering
    \includegraphics[width=1\linewidth]{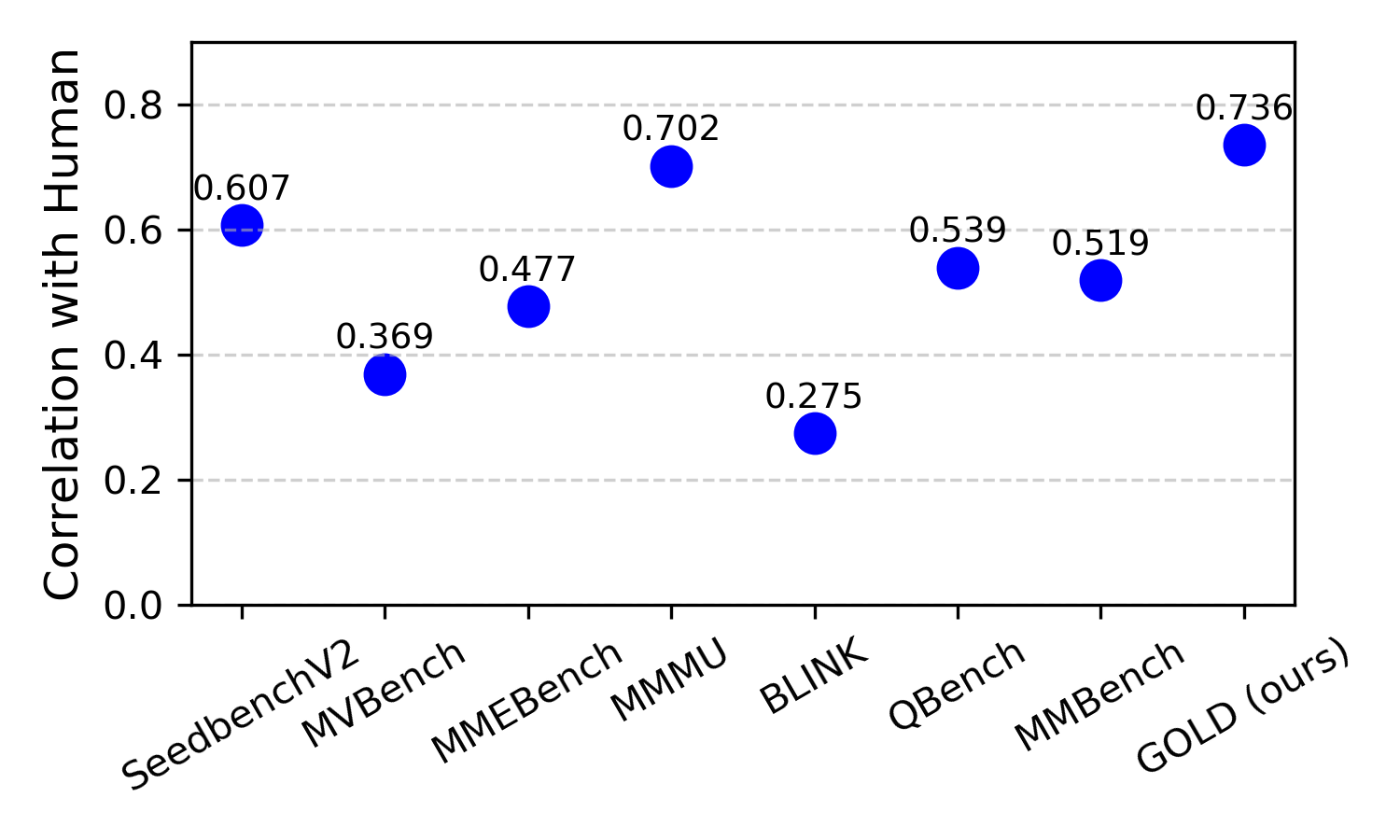}
    \caption{Correlation with human evaluation across benchmarks.}
    \label{fig:correlation_plot}
\end{figure}
\textbf{Correlation Analysis}
We evaluated the Pearson correlation between benchmark scores and human evaluation results, as shown in Fig.~\ref{fig:correlation_plot}. Among all compared benchmarks, \textsc{Gold} (ours) achieves the highest correlation (0.7359), demonstrating its superior alignment with human judgment. MMMU also shows strong consistency (0.702), whereas BLINK and MVBench exhibit weaker correlations (0.2746 and 0.369), suggesting that benchmarks may not effectively reflect real-world reasoning capabilities.

\section{Conclusion and Future Work}

We propose a statistics-theory-driven evaluation framework for MLLMs, grounded in Piaget’s cognitive development theory. By structuring model capabilities into perception, memory, and reasoning, we construct a new benchmark, \textsc{Gold}, to enable fine-grained, interpretable, and statistically valid evaluation. Experimental results show that \textsc{Gold} achieves better indicator validity, lower redundancy, and higher cognitive alignment compared to existing benchmarks, demonstrating the effectiveness of our framework.

Nevertheless, our approach is not without limitations. The current version of the \textsc{Gold} benchmark, though aligned with Piaget’s theory and effective in capturing core capabilities across perception, memory, and reasoning, may be overly simplistic for the latest generation of MLLMs. Many SOTA models already demonstrate near-perfect performance in several subtasks, leading to a potential ceiling effect that reduces the discriminative power of the benchmark. This also reflects the broader challenge of constructing cognitively grounded yet sufficiently challenging tasks that can meaningfully differentiate highly capable models. As part of future work, we plan to iteratively enrich the benchmark with more ambiguous and cognitively demanding items, better reflecting real-world task diversity and the advancing frontiers of MLLM capabilities.

\bibliography{IEEEabrv,ref}
\bibliographystyle{IEEEtran}

\section{Biography Section}
\vspace{-30pt}
\begin{IEEEbiography}[{\includegraphics[width=1in,height=1.25in,clip,keepaspectratio]{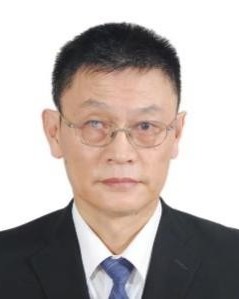}}]{Shengwu Xiong}
received the B.Sc. degree in computational mathematics from Wuhan University, Wuhan, China, in 1987, and the M.Sc. and Ph.D. degrees in computer software and theory from Wuhan University, in 1997 and 2003, respectively. Currently, he is the Dean of the Inter disciplinary Artificial Intelligence Research Institute, Wuhan College. He is also a Professor with the School of Computer Science and Artificial Intelligence, Wuhan University of Technology, China. His research interests include intelligent computing, machine learning, and pattern recognition.
\end{IEEEbiography}

\vspace{-30pt}
\begin{IEEEbiography}[{\includegraphics[width=1in,height=1.25in,clip,keepaspectratio]{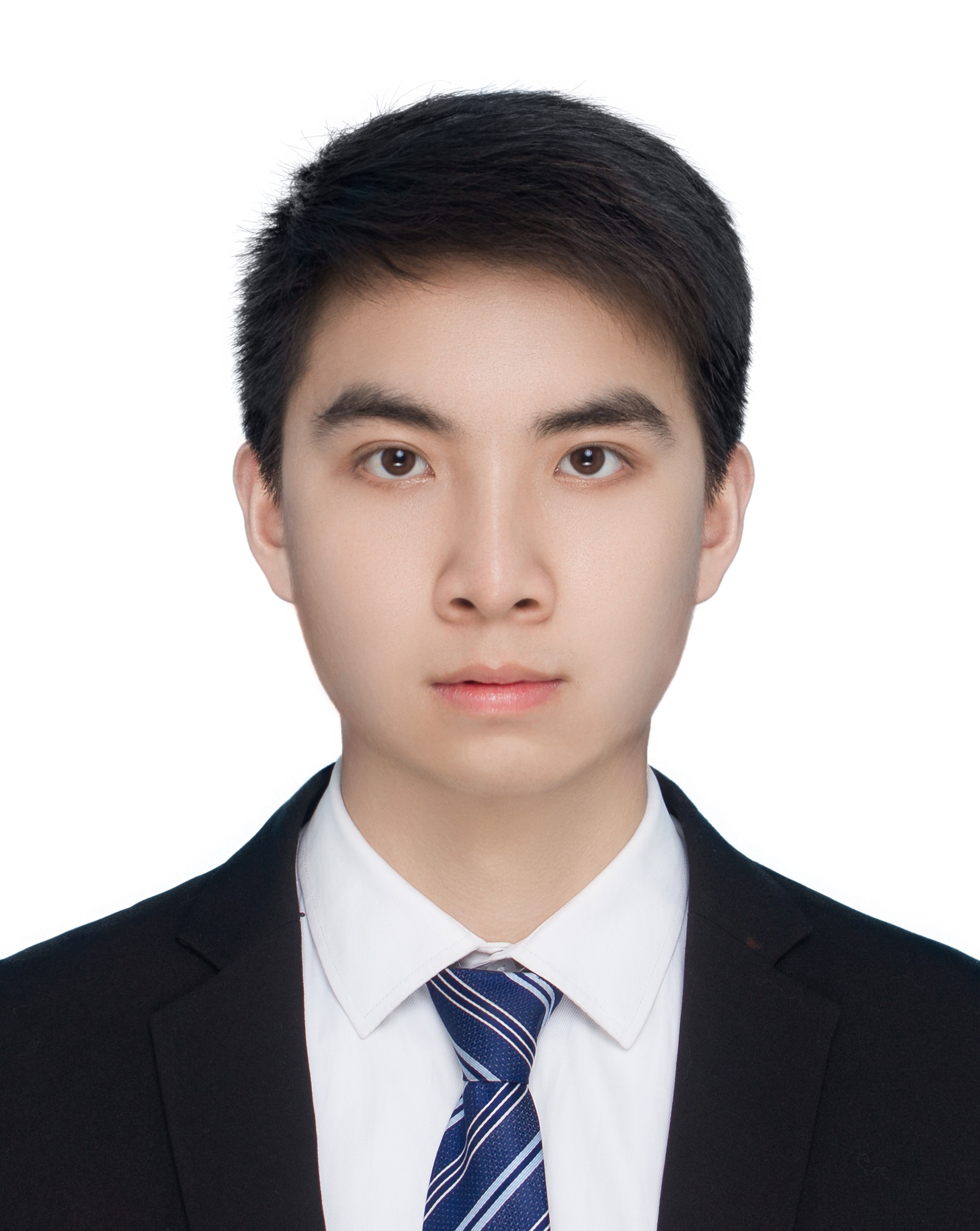}}]{Tianyu Zou}
received the B.Eng. degree from the Wuhan University of Technology, Wuhan, China, in 2022, where he is currently pursuing the Ph.D. degree with the School of Computer Science and Artificial Intelligence. His research interests include evaluation for multimodal large models, few-shot learning and semantic segmentation.
\end{IEEEbiography}

\vspace{-30pt}
\begin{IEEEbiography}[{\includegraphics[width=1in,height=1.25in,clip,keepaspectratio]{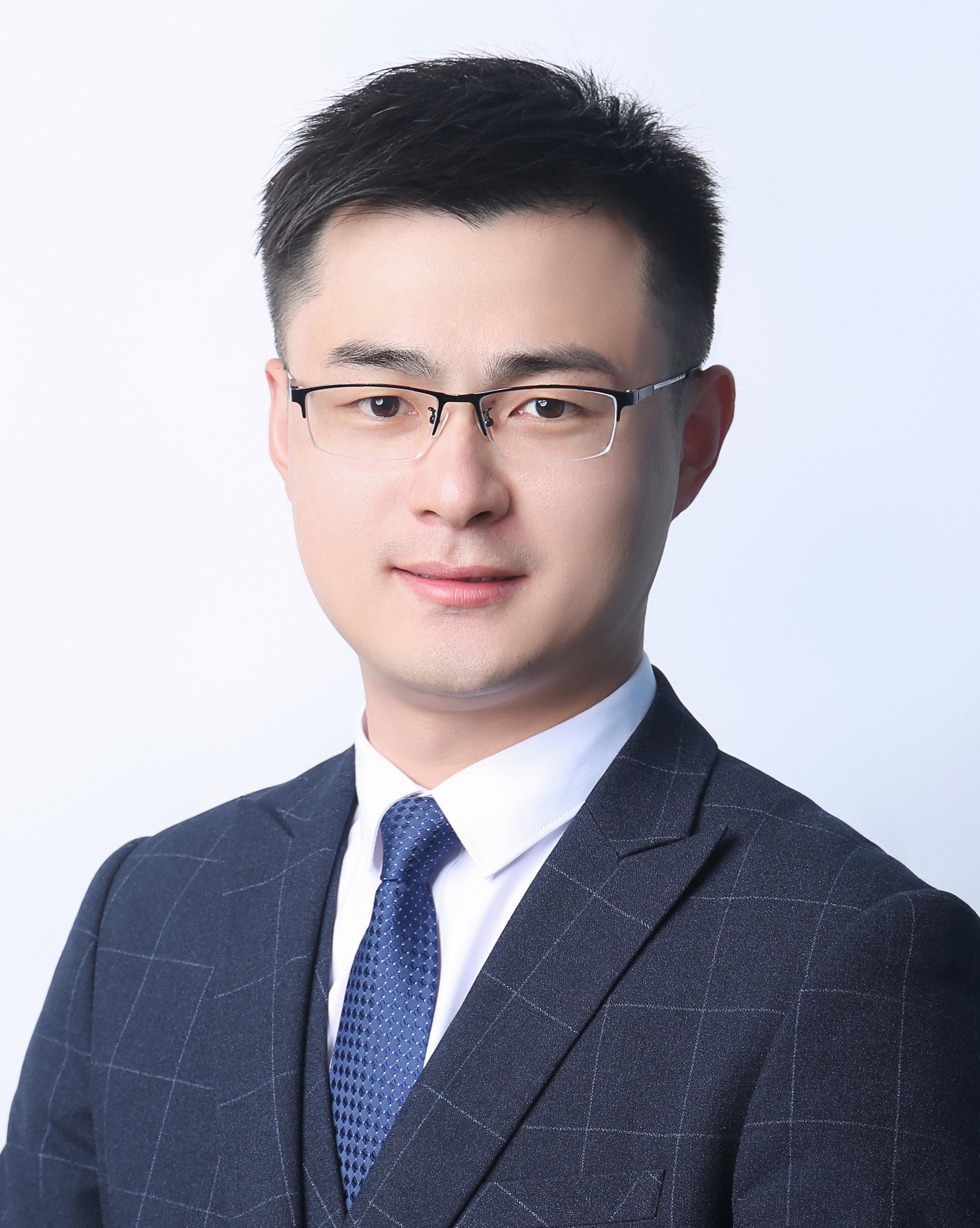}}]{Cong Wang}
(Member, IEEE) received the BS degree in automation and the MS degree in mathematics from Hohai University, Nanjing, China, in 2014 and 2017, respectively and the Ph.D. degree in mechatronic engineering from Xidian University, Xi’an, China, in 2021. 
He is currently an associate professor with the School of Mathematics and Statistics, Northwestern Polytechnical University, Xi’an. He was a Visiting Ph.D. Student with the Department of Electrical and Computer Engineering, University of Alberta, Edmonton, AB, Canada, and the Department of Electrical and Computer Engineering, National University of Singapore, Singapore. He was also a research assistant with the School of Computer Science and Engineering, Nanyang Technological University, Singapore. 
His current research interests include vicinagearth security, high-dimensional image analysis, and fuzzy theory and its applications. He currently has a textbook, a monograph, 10+ patents, and more than 40 papers (20+ in \textit{IEEE Transactions}) and hosts more than 10 research projects. 
He is funded by the China National Postdoctoral Program for Innovative Talents and the Excellent Chinese and Foreign Youth Exchange Program of the China Association for Science and Technology. He is an editorial board member of 10+ international journals like IEEE TFS and a program committee chair, track chair, publication chair, publicity chair, program committee member, and technical committee member of 60+ international conferences. He is also a frequent reviewer of 70+ international journals, including a number of the \textit{IEEE Transactions} and many international conferences.
\end{IEEEbiography}

\vspace{-30pt}
\begin{IEEEbiography}[{\includegraphics[width=1in,height=1.25in,clip,keepaspectratio]{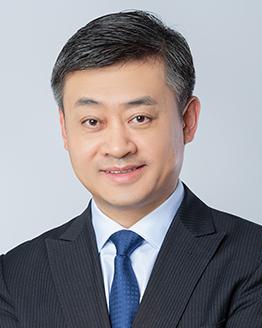}}]{Xuelong Li}
(Fellow, lEEE) is currently the CTO and chief scientist of China Telecom, where he founded the Institute of Artifcial Intelligence (TeleAl) of China Telecom.
\end{IEEEbiography}

\vfill

\end{document}